\documentclass[letterpaper]{article} 
\usepackage{aaai2026}  
\usepackage{times}  
\usepackage{helvet}  
\usepackage{courier}  
\usepackage[hyphens]{url}  
\usepackage{graphicx} 
\usepackage{natbib}  
\usepackage{caption} 
\usepackage{algorithm}
\usepackage{algorithmic}

\usepackage{newfloat}
\usepackage{listings}
\DeclareCaptionStyle{ruled}{labelfont=normalfont,labelsep=colon,strut=off} 
\floatstyle{ruled}
\newfloat{listing}{tb}{lst}{}
\floatname{listing}{Listing}
\floatstyle{ruled} 
\newfloat{prompt}{t}{lop}
\floatname{prompt}{Prompt}
\usepackage{multirow}
\usepackage{amsmath}
\usepackage{xcolor}

\usepackage{makecell}
\usepackage{subcaption}
 \usepackage{booktabs}
 \usepackage{graphicx}

\newcommand{\answerYes}[1]{\textcolor{blue}{#1}} 
\newcommand{\answerNo}[1]{\textcolor{teal}{#1}} 
\newcommand{\answerNA}[1]{\textcolor{gray}{#1}}

\title{MultiCaption: Detecting disinformation using multilingual visual claims}
\author{
Rafael Martins Frade\textsuperscript{1,2}\equalcontrib,
Rrubaa Panchendrarajan\textsuperscript{3}\equalcontrib,
Arkaitz Zubiaga\textsuperscript{3}
}
\affiliations{
    \textsuperscript{\rm 1}University of Santiago de Compostela, Spain \\
    \textsuperscript{\rm 2}Newtral Media Audiovisual, Spain \\
    \textsuperscript{\rm 2}Queen Mary University of London, United Kingdom \\
    mfrade.rafael@gmail.com, \{r.panchendrarajan, a.zubiaga\}@qmul.ac.uk
}

\usepackage{bibentry}

\begin{document}

\maketitle

\begin{abstract}
Online disinformation poses an escalating threat to society, driven increasingly by the rapid spread of misleading content across both multimedia and multilingual platforms. While automated fact-checking methods have advanced in recent years, their effectiveness remains constrained by the scarcity of datasets that reflect these real-world complexities. To address this gap, we first present \textit{MultiCaption}, a new dataset specifically designed for detecting contradictions in visual claims. Pairs of claims referring to the same image or video were labeled through multiple strategies to determine whether they contradict each other. The resulting dataset comprises 11,088 visual claims in 64 languages, offering a unique resource for building and evaluating misinformation-detection systems in truly multimodal and multilingual environments. We then provide comprehensive experiments using transformer-based architectures, natural language inference models, and large language models, establishing strong baselines for future research. The results show that \textit{MultiCaption} is more challenging than standard NLI tasks, requiring task-specific finetuning for strong performance. Moreover, the gains from multilingual training and testing highlight the dataset’s potential for building effective multilingual fact-checking pipelines without relying on machine translation.
\end{abstract}


\section{Introduction}
\begin{figure}
    \centering
    \includegraphics[width=1\linewidth]{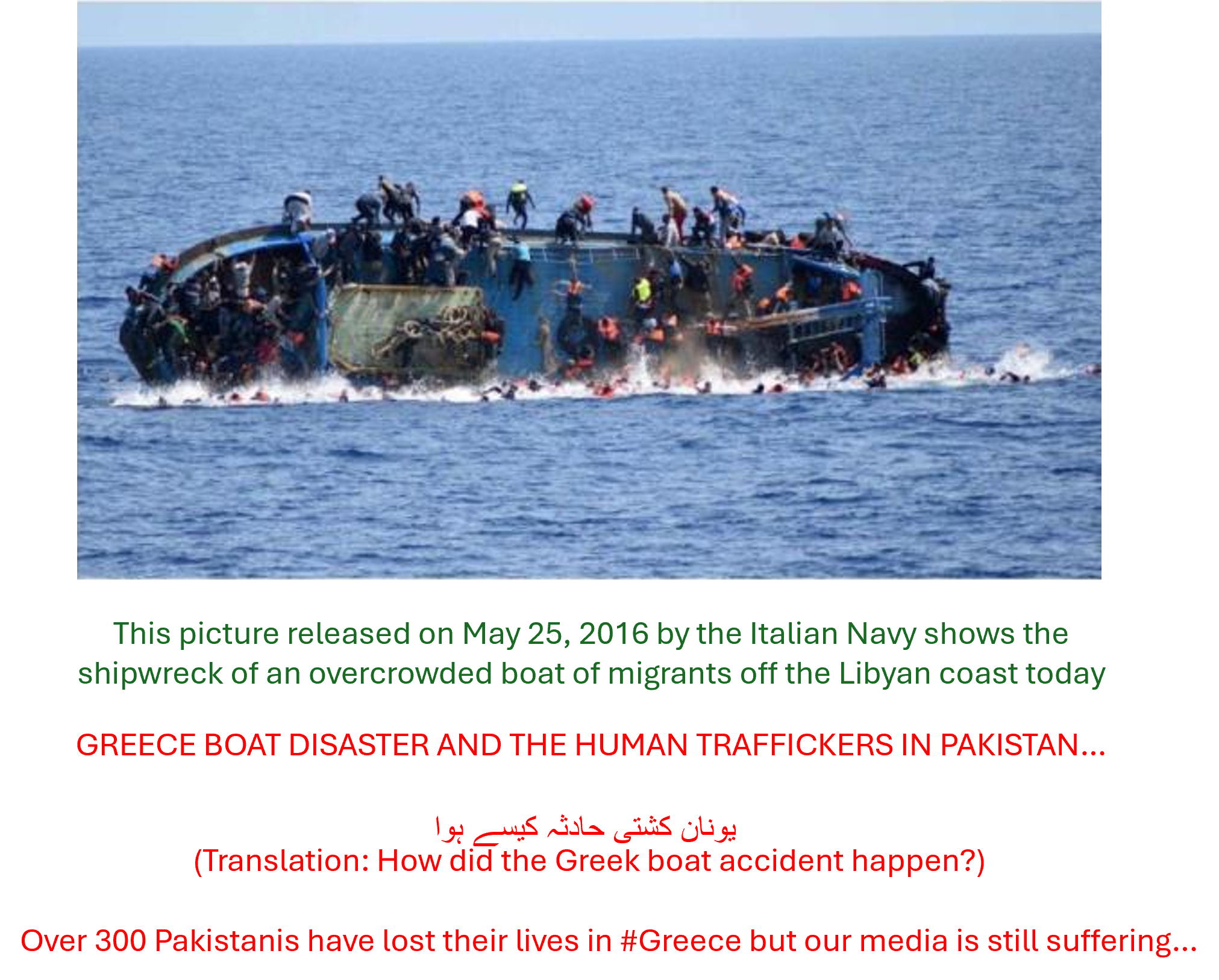}
    \caption{An example of image circulating in social media, with true (green) and false captions (red) \cite{afp_example}}
    \label{fig:contradiction_example}
\end{figure}

The rapid spread of misinformation online has its impacts identified in many aspects of society. Although automated fact-checking pipelines aim to mitigate this problem, the task remains highly challenging due to the complexity and diversity of misinformation. In particular, misinformation disseminated across multimedia content and multiple languages requires more generalizable and robust solutions. Figure \ref{fig:contradiction_example} illustrates this challenge: it shows an image with its original caption describing a 2016 incident, alongside false captions in English and Urdu that repurposed the same image in connection with an unrelated event in 2023. 

While early research in automated fact-checking focused primarily on textual claims \cite{ thorne2018fever, elsayed2019overview}, recent work has expanded to incorporate images, videos, and other multimodal and multilingual content \cite{akhtar2023multimodal, chakraborty2023information, aneja2023cosmos}. However, datasets specifically designed to address misinformation spread through images or videos remain scarce, and most existing resources focus on textual claim verification \cite{thorne2018fever, aly2021feverous}.


A closely related task aimed at identifying multimodal misinformation is out-of-context detection. Given an image and two associated claims, the objective is to determine whether one of the claims presents the image out of context. The dataset commonly used for this task is \textit{COSMOS} \cite{aneja2023cosmos}, which was primarily constructed from news articles and fact-checking websites. However, this dataset has certain limitations: it contains only claims in English and includes a relatively small number of instances that reflect claims used to spread disinformation in real-world scenarios.

To fill this gap, we first introduce \textit{MultiCaption}, a multilingual dataset designed to identify disinformation through contradictory visual claims. We define two visual claims as contradictory if they cannot both be true simultaneously about the same image or video  \cite{sepulveda2023detecting}. We build our dataset on \textit{MultiClaim} \cite{pikuliak2023multilingual} and \textit{MultiClaimNet} \cite{multiclaimnet} as primary data sources, which consists of claims written by professional fact-checkers and based on social media content used to disseminate real-world misinformation across multiple countries and languages. We employ multiple labeling strategies, including manual validation, annotation using large language models (LLMs) and methods leveraging claim links from the original sources to label visual claim pairs as contradictory or non-contradictory. We apply a rigorous filtration process to extract high-quality visual claim pairs from the sources, constructing a dataset that reflects real-world challenges. This process yields 11,088 claim pairs across 64 languages, encompassing both monolingual and cross-lingual pairs, making \textit{MultiCaption} a valuable resource for advancing research in multilingual disinformation detection.


Using the \textit{MultiCaption} dataset, we then perform extensive experiments, carefully constructing a disjoint test set to prevent data leakage and applying data expansion strategies on the training set to enable robust fine-tuning. We provide benchmark results from fine-tuned transformers, natural language inference (NLI) models, and LLMs on both the \textit{MultiCaption} test set and \textit{COSMOS}, in monolingual and multilingual settings. Our results show that \textit{MultiCaption} introduces a more challenging benchmark than \textit{COSMOS}. Moreover, most models benefit significantly from the fine-tuning process, particularly in the multilingual setting, highlighting the value of our dataset as a resource for advancing multilingual misinformation detection.

We make the following contributions:
\begin{itemize}
    \item We introduce \textit{MultiCaption}\footnote{Dataset is available at \url{https://doi.org/10.5281/zenodo.18230659}, and source code is available at \url{https://github.com/rfrade/multicaption}}, the first multilingual dataset for detecting contradictory visual claims, containing 11,088 claim pairs across 64 languages.
    \item We carefully curate multiple labeling strategies, including both manual and LLM-based methods, to label visual claim pairs as contradictory or non-contradictory.
    \item We conduct extensive experiments to establish robust baselines in a multilingual setting and explore strategies for further expanding training data.
\end{itemize}

We believe our dataset and experiments will serve as a valuable resource and benchmarking for developing effective contradiction detection systems to combat disinformation in multilingual settings. 

\section{Related Work}
\label{related_work}
We provide an overview of existing research and datasets relevant to contradiction detection, as well as closely related tasks such as natural language inference (NLI), out-of-context detection, and claim matching.

\paragraph{\textbf{Contradiction Detection \& NLI}}
Identifying contradictions between two pieces of text is typically framed as a NLI task \cite{li2017contradiction}. More fine-grained task definitions address contradiction-type classification \cite{senouci2025claim}, detecting contradictions between a query and a document \cite{xu2024sparsecl}, identifying contradictions within a single document \cite{li2024contradoc}, and cross-modal contradictions between images and text \cite{popordanoska2025clash}. 

The NLI task involves identifying entailment, neutrality or contradiction between two sentences \cite{gubelmann2024capturing}. NLI techniques are widely used in disinformation research, especially for claim verification. In this context, labels are often adapted from the standard NLI schema (CONTRADICT, ENTAIL, NEUTRAL) to task-specific ones such as SUPPORTED, REFUTED, and NOT ENOUGH INFO \cite{thorne2018fever}. As in many other natural language processing tasks, claim verification has recently shifted toward leveraging LLMs \cite{ahmad2025climatecheck, dmonte2024claim}.
Two common datasets used in this task are FEVER \cite{thorne2018fever} and FEVEROUS \cite{aly2021feverous}. Both contain claims accompanied by textual and tabular evidence that supports or refutes each claim. A limitation of those datasets is that the data is only in English and was generated based on Wikipedia, lacking semantic characteristics commonly present on social media language.



\paragraph{\textbf{Out-of-context (OOC) Detection}} Despite growing focus on deepfakes, a common form of disinformation involves real images or videos misrepresented through false context or claims. Closely related to the automated fact-checking pipeline, OOC detection evaluates the truthfulness of a claim given an image or video. The task typically assumes access to the original context as a triplet \textit{(Candidate Claim, Original Claim, Image)} and aims to determine entailment or contradiction between the claims, optionally incorporating visual information. Most work in this setting relies on the \textit{COSMOS} dataset and explores approaches such as context retrieval, text similarity, contradiction detection, entity linking, LLMs, and cross-modal consistency \cite{aneja2023cosmos, abdelnabi2022open, nguyen2023multi, la2022combination, tran2022textual, nguyen2022grit, la2022leverage}.

Despite introducing the task and shaping research on the topic, the COSMOS dataset has two main limitations: It contains only claims in English and includes a very small labeled dataset (1,700 samples), containing only a limited number of real disinformation cases. Another limitation is that its OOC class contains a substantial amount of direct negations or linguistic features in candidate claims suggesting that they are fact-checks, making it less useful to develop solutions for real-life disinformation detection. 

Another line of work assumes that the original context is unavailable and attempts to determine whether an image or video is presented out of context using only the claim–image pair. This approach relies on datasets such as NewsClipping \cite{luo2021newsclippings} and VERITE \cite{papadopoulos2024verite}, and typically applies multimodal feature extraction and classification \cite{papadopoulos2025similarity}. However, in many real-world social media scenarios, the veracity of accompanying claims cannot be reliably assessed without access to the original context; Figure \ref{fig:contradiction_example} shows an example illustrating this limitation.

\begin{figure*}[t]
    \centering
    \includegraphics[width=0.9\linewidth]{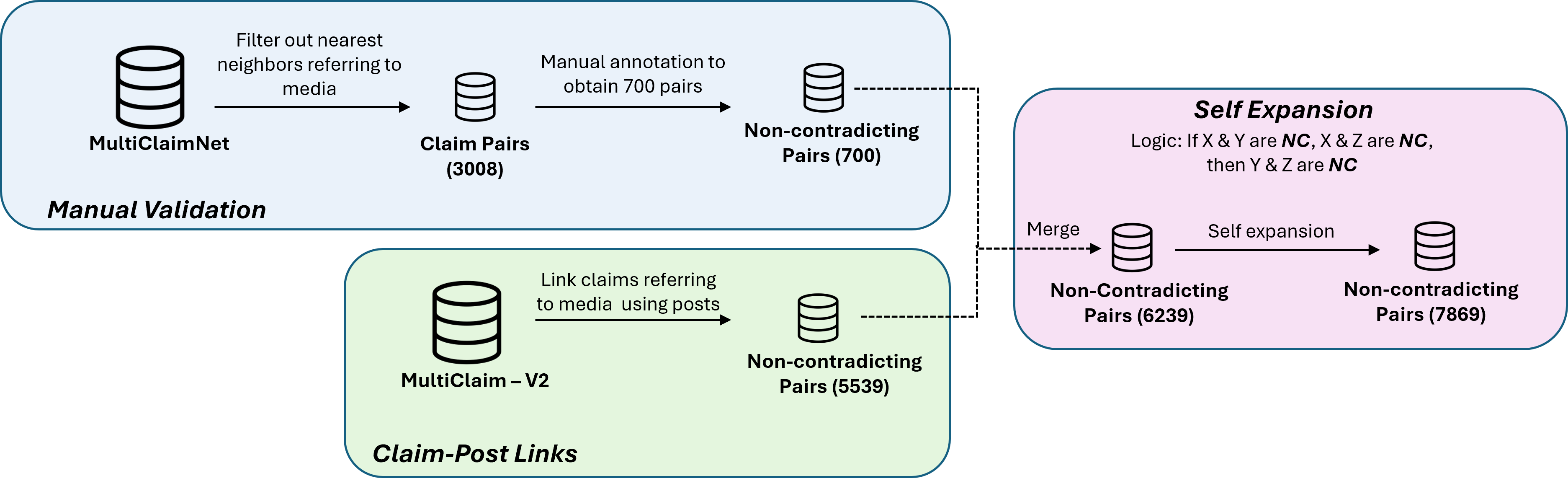}
    \caption{Work flow of generating non-contradicting (NC) pairs}
    \label{fig:work_flow_nc}
\end{figure*}

\paragraph{\textbf{Claim Matching}}
Claim matching is a fundamental component of automated fact-checking pipelines. The task involves determining whether two claims convey similar information and can be extended to matching an unverified claim against a database of verified claims, a process often referred to as fact-checked claim retrieval \cite{panchendrarajan2024claim}. While claim matching is framed as a binary task—deciding whether two claims are similar or not \cite{larraz2023semantic}—non-similarity does not necessarily imply contradiction. Claim matching typically focuses on fine-tuning multilingual transformers such as XLM-R \cite{larraz2023semantic} or LLMs  \cite{pisarevskaya2025zero}. While several claim matching datasets exist \cite{nielsen2022mumin,singh2023finding,pikuliak2023multilingual}, they are designed for similarity detection and cannot be used to detect contradictions without significant adaptation. 


\section{MultiCaption Dataset Construction}
\label{dataset_construction}
\paragraph{Definition of Contradicting Visual Claims:} We define a pair of claims referring to the same image or video (referred to as visual claims) as \textit{contradictory} if both cannot be true at the same time \cite{sepulveda2023detecting}. A claim being less specific or only partially correct does not constitute a contradiction.

\paragraph{Source of Visual Claims:} We use the \textit{MultiClaim} \cite{pikuliak2023multilingual}  dataset as the main source of multilingual claims referring images or videos for constructing the contradicting and non-contradicting pairs. The dataset comprises fact-checked claims along with  references to the fact-checking articles and social media posts discussing these claims. The dataset was released in two versions (\textit{MultiClaim} v1 and v2), with the final version containing 435K fact-checked claims and 89K linked social media posts. Each social media post is linked to at least one claim, resulting in a total of 105K claim–post links. The initial stages of our dataset construction were based on \textit{MultiClaim} v1, while later stages used \textit{MultiClaim} v2.  We used the terms ["picture", "image", "photo", "photograph", "footage", "document", "video", "clip", "post"] combined with the verb "show" (e.g., photo shows) or preposition "of" (e.g., video of) present in the English translation of the text to obtain the visual claims.

We employ a combination of manual and automatic methods to label candidate claim pairs, with the workflow for generating contradicting and non-contradicting pairs detailed below.

\subsection{Generating Non-Contradicting Pairs}
We employ three labeling strategies to generate non-contradicting pairs—manual validation, claim–post links, and self-expansion—as shown in Figure~\ref{fig:work_flow_nc}.

\subsubsection{Generation using Manual Validation}\label{sec:manual_nonc}
\begin{figure*}[t]
    \centering
    \includegraphics[width=0.9\linewidth]{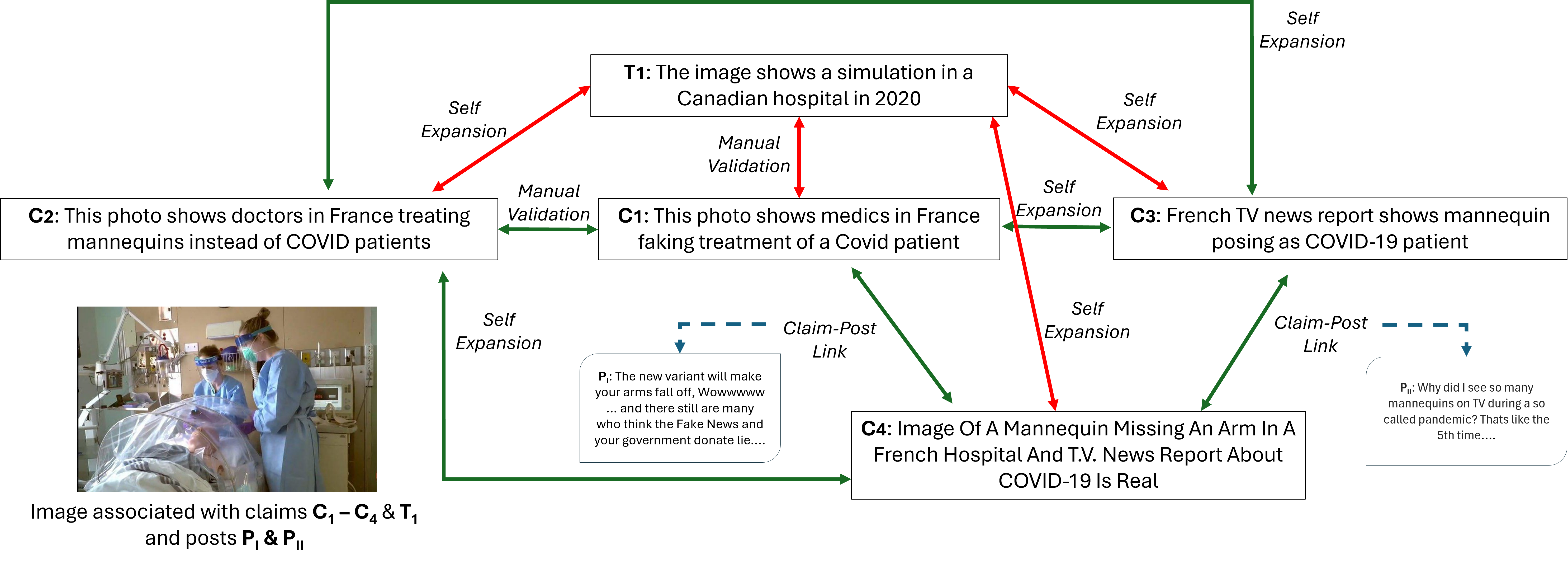}
    \caption{Example illustrating different annotation strategies in MultiCaption. Arrow colors encode relationship types: green for non-contradicting, red for contradicting, and blue for associated posts}
    \label{fig:example}
\end{figure*}

Manual validation requires curating a set of visual claim pairs that are most likely to be non-contradicting to each other. We use \textit{MultiClaimNet} \cite{multiclaimnet}, derived from \textit{MultiClaim} v1, which contains claim pairs constructed using a nearest-neighbor approach and annotated for similarity with large language models (LLMs). In the original work, the authors further grouped similar claims into clusters. From \textit{MultiClaimNet}, we filtered out pairs in which at least one claim refers to an image or video, resulting in 3,008 similar claim pairs. However, as the authors relied solely on textual information for similarity annotation, these pairs do not necessarily refer to the same image or video. Therefore, our manual verification involved checking whether both claims refer to the same image or video, and ensuring that the claims are not contradictory. The original media was obtained from the fact-checking articles linked to each claim. Figure \ref{fig:example} shows an example of manually validated claim pairs $(C_1, C_2)$. 

Since the claim pairs in \textit{MultiClaimNet} were generated using a nearest-neighbor approach, we observed a similarity bias in the sample, where most pairs exhibit high similarity (ranging from 0.52 to 1) with an average of 0.86. Claim similarity was measured using cosine similarity between embeddings of their English translations, generated with Sentence Transformer \cite{reimers-2019-sentence-bert}. To mitigate this bias, we verified claim pairs in order of increasing similarity—starting from the least similar—until we obtained 700 valid non-contradicting pairs. The number of valid pairs was selected to closely match the number of manually validated contradicting pairs. To reach this target, we manually validated 1,144 claim pairs from the original sample. A substantial portion (approximately 33\%) was discarded because the claims referred to different media, while an additional 5\% were excluded due to missing fact-checking articles. Only a small fraction of pairs (less than 1\%) were found to be contradictory within the \textit{MultiClaimNet} sample.

\subsubsection{Generation using Claim-Post Links}
\textit{MultiClaim} dataset comprises claims and social media posts linked if the post discusses the claim. This relationship between claims and posts can be used to link similar claims referring to the same post. Therefore, we create further non-contradicting pairs when two visual claims share a post. Figure \ref{fig:example} shows claim pairs $(C_1,C_4)$ and $(C_3,C_4)$ automatically labeled non-contradicting via posts $P_I$ and $P_{II}$ respectively. 

The latest version of \textit{MultiClaim} comprises 105K claim–post links, created using multiple strategies, including back-linking, claim review schemas, and identical claims \cite{pikuliak2023multilingual}. We excluded links generated through identical claims, as they inherently produce obvious non-contradicting pairs. Further, linking two claims via a post requires the post to be associated with at least two claims. Applying this constraint yielded 10.6K potential links, corresponding to 16.2K claim pairs. We retained only the pairs in which at least one claim referred to an image or video, resulting in 6,182 non-contradicting claim pairs.

Since these pairs were generated automatically, we manually inspected samples from both the lowest- and highest-similarity ranges at 0.05 similarity intervals.  During this analysis, we observed that pairs with very low similarity scores ($< 0.4$) were predominantly noisy, while those with very high similarity scores ($> 0.95$) often contained identical claims. To eliminate noise and trivial examples, we discarded claim pairs from both extremes of the similarity distribution. Highly similar pairs ($> 0.95$) were removed only when both claims were in the same language. Multilingual pairs were retained, as they contribute valuable cross-lingual context to the dataset despite their high similarity. Refer to the Appendix for the similarity distribution of the claim pairs created via claim–post links and the discarded regions.

\subsubsection{Generation using Self-Expansion}
Similar to the claim–post links discussed previously, there also exist direct links between claims. We identified that if claim $X$ and claim $Y$ are labeled as non-contradicting, and claim $X$ and claim $Z$ are also labeled as non-contradicting, then claim $Y$ and $Z$ can be automatically inferred to be non-contradicting. We refer to this process as self-expansion, through which additional non-contradicting pairs are automatically generated. This process yielded 1,768 new claim pairs labeled as non-contradicting via self-expansion. Figure \ref{fig:example} illustrates examples, including $(C_1, C_3)$ (via $C_4$), $(C_2, C_3)$, and $(C_2, C_4)$ (via $C_1$), all automatically labeled non-contradicting.

Together, the three labeling techniques—manual validation, claim–post linking, and self-expansion—produced a total of 7,869 non-contradicting claim pairs, as summarized in Table \ref{tab:label_stats}.

\begin{table}[!t]
\centering
\footnotesize
\begin{tabular}{llr}
\hline
\textbf{Label} & \textbf{Labeling Strategy} & \textbf{\# Pairs} \\ \hline
\multirow{4}{*}{Contradiction} 
 & Manual Validation & 722 \\
 & LLM Annotation & 2072 \\
 & Self-Expansion & 425 \\
 & \textbf{Total} & \textbf{3219} \\ \hline
\multirow{3}{*}{Non-Contradiction} 
 & Manual Validation & 700 \\
 & Claim-Pair Link & 5508 \\
 & Self-Expansion & 1661 \\ 
 & \textbf{Total} & \textbf{7869} \\ \hline
\end{tabular}
\caption{Labeling strategies and No. of pairs in \textit{MultiCaption}.}\label{tab:label_stats}
\end{table}

\begin{figure*}
    \centering
    \includegraphics[width=0.9\linewidth]{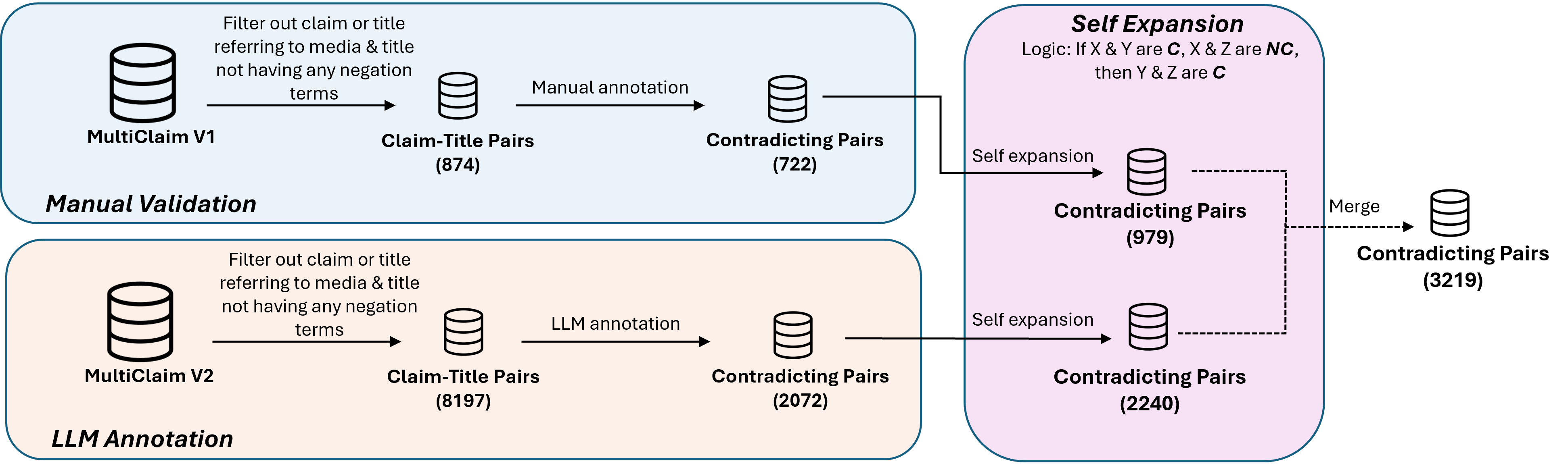}
    \caption{Work flow of generating contradicting (C) pairs}
    \label{fig:work_flow_c}
\end{figure*}

\subsection{Generating Contradicting Pairs}

As mentioned earlier, \textit{MultiClaim} contains claims and fact-checking articles written by professional fact-checkers. Since every claim is associated with at least one fact-checking article, we used the title of the fact-checking article as a source of contradicting claims. Similar to non-contradicting pairs, we generate contradicting pairs using various labeling strategies—including manual validation, LLM annotation, and self-expansion as shown in Figure~\ref{fig:work_flow_c}.

Fact-checking article titles typically follow three common patterns: (1) rewriting the claim as a question, (2) explicitly negating the claim, or (3) providing the original context of an image or video. The first pattern is not directly useful for our purposes, and the second—an explicit negation—would be trivially detectable in an automated fact-checking pipeline. We therefore focus on the third pattern, in which the title reveals the true context of the referenced image or video. The filtration aimed to create a dataset simulating real-world fact-checking, given an original context of an image or video, an automated agent determines whether a claim presents it out of context.

\subsubsection{Generation using Manual Validation}\label{sec:con_manual}

Based on claim-title pairs from \textit{MultiClaim} v1, we first retained only the pairs in which either the claim or the title referred to a media item. We then removed titles that contained explicit negations or direct references to the claim. For example, for a claim such as "Image of the protest in France 2025", we excluded titles like "This image is not from France" or "Image of protest actually dates back to 2020". The exclusion of such cases represents a key advantage of our dataset relative to COSMOS. Refer to Table \ref{tab:negation_terms} in Appendix for negation terms used to filter out title with direct reference or denial of the claim. We then manually validated this sample to retain only the titles that did not contain any additional form of negation and did not make a direct reference to the claim, resulting in 722 contradicting claim–title pairs. The objective was to end up with pairs in which both claim and title were independent claims about the image/video. Figure \ref{fig:example} shows an example of manually validated claim-title pair $(C_1, T_1)$. 


\subsubsection{Generation using Self-Expansion}\label{sec:con_self_expansion} 
Similar to the self-expansion technique used to generate non-contradicting claim pairs, the inherent relationships between contradicting and non-contradicting claims can be leveraged to automatically generate additional contradicting samples. Specifically, we also identified that if claim X contradicts claim Y, and claim X is non-contradicting with claim Z, then claim Y and claim Z are also contradicting. Figure \ref{fig:example} illustrates an example of this expansion: self-expanded contradicting pairs $(T_1, C2)$, $(T_1, C3)$, and $(T_1, C4)$ are generated as $(T_1, C1)$ is manually labeled as contradicting and $C_1$ is non-contradicting with $C_2 - C4$. Using this method, 722 manually validated contradicting pairs were expanded to a total of 979 pairs.

\subsubsection{Generation using LLM-Annotation}
The number of contradictory pairs obtained through manual validation and self-expansion was relatively small compared with the non-contradictory pairs. Nevertheless, the \textit{MultiClaim} v2  contains many potential contradicting claim–title pairs. Since manually validating all candidate pairs is infeasible, we leverage the LLM GPT-5 to perform validation as a proxy for manual review. 

We applied the same filtering process described in Section \ref{sec:con_manual} to extract potential contradicting claim–title pairs from the \textit{MultiClaim} v2 dataset. This procedure resulted in 8,197 claim–title pairs. For LLM annotation, we constructed the prompt to closely mimic the manual validation process. Although a claim and a title may appear contradictory, some titles directly reference the claim while explicitly denying it. To handle this, we instructed the LLM to label each claim–title pair as: \textit{contradict} if both cannot be true for the same image/video, and \textit{denial} if the title explicitly refutes or debunks, even with added context. Only claim–title pairs that are labeled as \textit{contradict} but not \textit{denial} are considered valid contradictory pairs. 

To reduce potential noise from LLM annotations, we evaluated the approach using 874 claim–title pairs that were manually validated and labeled as either valid or invalid contradictory pairs. We iteratively refined the annotation prompt with clearer instructions and illustrative examples to guide accurate predictions. Once the prompt is finalized, we re-labeled the same 874 manually validated pairs using LLM to quantitatively assess its precision in identifying valid contradicting claim–title pairs. We measured the precision in deciding a pair as valid (\textit{contradict}=True, \textit{denial}=False) to limit potential noise introduced through LLM-based annotation. The final prompt yielded a precision of 0.887. This prompt was then applied to annotate the 8,197 candidate claim–title pairs. Refer to Appendix \ref{appendix:llm_annotation} for the prompt used for LLM annotation and the distribution of \textit{contradict} and \textit{denial} labels. Consequently, 2,072 pairs (around 25\%) were retained as valid contradicting claim–title pairs. These pairs were further expanded using the self-expansion technique and merged with manually validated pairs, resulting in a total of 3,219 contradicting pairs.

\begin{figure}
    \centering
\includegraphics[width=1\linewidth]{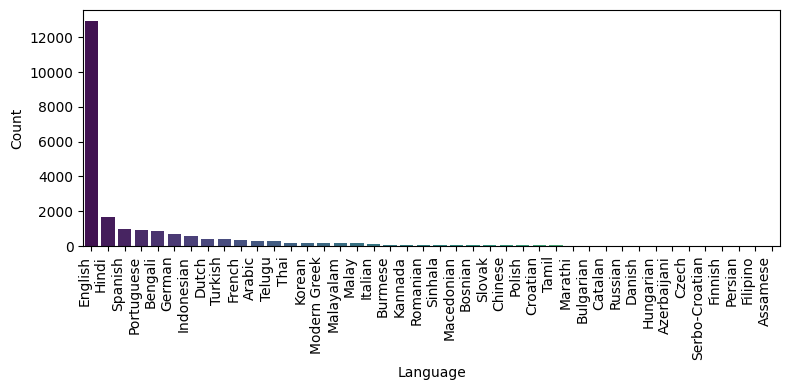}
    \caption{Language distribution of \textit{MultiCaption}}
\label{fig:language_distribution}
\end{figure}

\subsection{Multilingual Claims}
The original source, \textit{MultiClaim}, is predominantly multilingual; consequently, \textit{MultiCaption} also comprises claims written in 64 languages. Figure \ref{fig:language_distribution} illustrates the distribution of languages appearing at least 5 times in the dataset. Among these, English is the most frequent, followed by Hindi, Spanish, Portuguese, and Bengali. Among the 11K claim pairs, 8,131 are monolingual and 2,957 are multilingual, hence enabling multi- and cross-lingual research. Within the multilingual subset, Hindi–English combination constitutes the largest portion, with 973 pairs.

\subsection{Claim Topics}

Table \ref{tab:topics} presents the top 20 topics identified with BERTopic \cite{grootendorst2022bertopic} clustering. As expected, many reflect major events from recent years. Topics with global relevance tend to appear across a larger number of languages. Other topics are more geographically specific, but strong diffusion within particular regions may explain their high volume of associated claims.

\section{Experiment Setup}
\label{experiment_setup}
\subsection{Datasets}
In addition to our \textit{MultiCaption} dataset that we introduce, we evaluate several baselines on the widely used out-of-context benchmark \textit{COSMOS} \cite{aneja2023cosmos} to enable broader comparison. \textit{COSMOS} contains 1,700 English image captions labeled as either out-of-context or not. We partition \textit{MultiCaption} into training and test splits, fine-tune baseline models on the training split, and report performance on both the \textit{MultiCaption} test split and \textit{COSMOS}. To assess the multilingual capabilities of the baselines, we experiment with two language configurations in \textit{MultiCaption}:
\begin{itemize}
    \item Monolingual - English translations of the claims (obtained from \textit{MultiClaim}) are used for both training and testing. Performance on the \textit{COSMOS} dataset is evaluated using these monolingual models.
    \item Multilingual - The original multilingual claims are used for both training and testing. 
\end{itemize}
The following subsection details the procedure used to construct the \textit{MultiCaption} train/test split.

\begin{table}[t]
\centering
\footnotesize
\label{tab:topics}
\begin{tabular}{lcc}
\toprule
\textbf{Topic Description} & \textbf{\# Claims} & \textbf{\# Languages} \\
\midrule
Russian invasion of Ukraine & 348 & 28 \\
Riots in France & 177 & 17 \\
COVID-19 vaccination & 137 & 19 \\
Accident in Kerala & 133 & 18 \\
Pandemic & 123 & 21 \\
UFO-related news & 116 & 12 \\
Gaza fake scenes & 109 & 17 \\
Military clashes in India & 108 & 11 \\
Prime Minister Modi & 99 & 6 \\
Wildfires & 90 & 15 \\
South Korean politics & 90 & 11 \\
Rescue operations in Turkey & 88 & 23 \\
Murder of Hindu girl & 87 & 8 \\
Donald Trump & 85 & 11 \\
Farmers’ protests & 84 & 11 \\
Ancient discoveries in Ayodhya & 81 & 14 \\
Joe Biden/Hunter Biden case & 81 & 7 \\
Conflict in Myanmar & 77 & 8 \\
Child kidnapping and trafficking & 76 & 16 \\
Israeli soldiers & 74 & 14 \\
\bottomrule
\end{tabular}
\caption{Top 20 topics in \textit{MultiCaption}}
\label{tab:topics}
\end{table}

\subsubsection{Test Partition Split}
\label{test_partition_split}
The goal of constructing a test partition was to create a challenging set of visual claim pairs for evaluation—one that is balanced across classes, reflects a diverse distribution of textual similarities, and remains fully disjoint from the training set. We initially considered the 1,422 manually validated pairs as the test set. However, this partition includes many highly similar non-contradicting pairs due to the source as discussed earlier. Further, due to the multiple annotation strategies, the same claim may appear across different samples. In the worst case, a random split could allow relational leakage: for example, if pairs $(A,B)$ and $(A,C)$ appear in training and $(B,C)$ appears in testing, the test set would no longer be independent. It is therefore essential to construct two strictly disjoint sets such that no claim—and no pairwise relationship between claims—present in the test set appears in the training set. We therefore apply an iterative procedure to create a disjoint and balanced test partition as follows:
\begin{enumerate}
    \item Initialize the test set with the 1,422 manually validated pairs.
    \item Enforce disjointness: move any pair from the training set to the test set if it contains a claim already present in the test set. For example, if $(B,C)$ appears in the test set, then $(A,B)$ and $(A,C)$ are moved from training to test.
    \item Balance classes within similarity bins: for each similarity bin of width 0.05 between 0 and 1, randomly sample instances from training and move to the test set to equalize the number of contradictory and non-contradictory pairs whenever possible.
    \item Repeat step 2 \& 3 until no additional samples can be moved from training to test.
    \item Finalize class balance: within each similarity bin, randomly discard samples from the majority class so that both classes contain equal counts in test set. Manually annotated samples were not discarded at this step.
\end{enumerate}

\begin{figure}
    \centering
\includegraphics[width=0.75\linewidth]{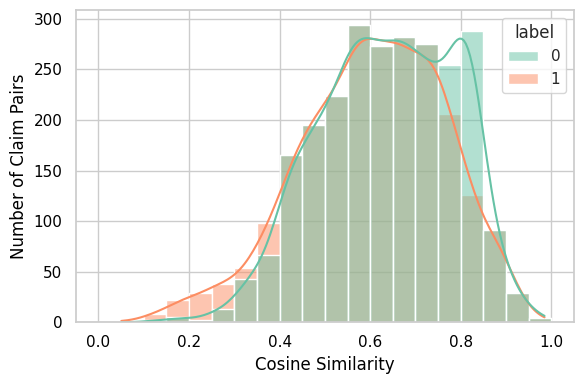}
    \caption{Textual similarity distribution in \textit{MultiCaption} - test parition}
\label{fig:similarity_distribution}
\end{figure}

The iterative process produced a test partition of 4920 claim pairs spanning 52 languages, with an approximately equal number of contradicting and non-contradicting pairs and a balanced similarity distribution, as shown in \ref{fig:similarity_distribution}. Minor imbalances within certain bins arise from retaining the manually validated pairs during Step 5.

\subsubsection{Training Data Expansion}
Creating a highly balanced test partition resulted in highly imbalanced train set, containing 4747 non-contradicting pairs and only 804 contradicting pairs. Since our primary goal was to establish a robust benchmark for identifying contradictory visual claims, we prioritized the integrity of the test set.

To mitigate the imbalance of contradicting pairs in training, we adopted a training data expansion strategy using GPT-5. For the 804 contradicting pairs, we instructed the model to generate a paraphrase of the claim and a paraphrase of the title in English to avoid noise (refer to Appendix \ref{sec:appendix_data_expansion} for more details and prompt used). Each contradicting pair (A,B) produced two additional pairs: (A, Paraphrase B) and (Paraphrase A, B), yielding 1,608 additional training pairs.

Our preliminary results revealed that some baselines trained in the \textit{MultiCaption} training partition performed poorly on the \textit{COSMOS} dataset. One contributing factor is that 56\% of \textit{COSMOS} contradicting examples (out-of-context pairs) contains one of the negation expressions we initially used to filter out claims from \textit{MultiCaption}. To train models capable of detecting both types of contradiction—independent claims and negations—we therefore further expanded the training set with \textit{MultiClaim} claim-title pairs in which the title is a direct negation of the claim. A similar set of negation expressions\footnote{We excluded certain expressions that frequently appear in titles in negation form but may not indicate true negation: ``claim that'', ``nothing'', ``generated'', ``fact check:'', ``associated with'', ``context'', ``since'', ``not'', ``actually'', ``as if'', ``yes'', ``in reports about'', ``was taken in'', ``attention'', ``claiming''} used to filter out claim-title pairs initially (Table \ref{tab:negation_terms}) was used to select pairs to be included in the train set. Following the paraphrasing expansion strategy, we added 1,608 direct negation pairs. We also observed that non-contradicting pairs tend to be more similar than contradicting pairs. To decrease the imbalance, the 1,608 negation pairs were randomly selected from the high-similarity region (cosine similarity 0.5–0.85). The resulting expanded training set contains 4,020 contradicting pairs and 4,767 non-contradicting pairs across 59 languages.

Table \ref{tab:dataset_stats} summarizes the statistics of the \textit{MultiCaption} train and test partitions, as well as \textit{COSMOS}.

\begin{table}[!t]
\centering
\footnotesize
\begin{tabular}{lcccc}
\hline
\textbf{Dataset} &
\rotatebox{90}{\textbf{\# C Pairs}} &
\rotatebox{90}{\textbf{\# NC Pairs}} &
\rotatebox{90}{\textbf{Total Pairs}} &
\rotatebox{90}{\textbf{\# Languages}} \\ \hline
MultiCaption - Train & 4020 & 4767 & 8795 & 59  \\
MultiCaption - Test  & 2415 & 2505 & 4920 & 52  \\
COSMOS              & 850  & 850  & 1700 & English \\ \hline
\end{tabular}
\caption{Statistics of datasets used in the experiments}
\label{tab:dataset_stats}
\end{table}

\subsection{Baselines}

We employ the following three types of models to establish stronger baselines for detecting contradictory visual claims. Refer to Appendix \ref{appendix:baselines} for the implementation of baselines.

\paragraph{Transformer-based Classifiers.} We fine-tuned the following three multilingual transformer architectures for sentence classification: XLM-Roberta-large (XLM-R) \cite{conneau2020unsupervised}, Multilingual DeBERTa-V3 (mDeBERTa) \cite{he2021debertav3}, and Multilingual BERT (mBERT) \cite{devlin2019bert}.

\paragraph{Natural Language Inference (NLI) Models.} Since our task is closely related to NLI, we also evaluated multilingual NLI models. These models are transformer encoders equipped with a classification head fine-tuned on NLI datasets to predict one of three labels— entailment, neutral, or contradiction—for a given sentence pair. We fine-tuned the NLI models XLM-RoBERTa-large-xnli \cite{alotaibi2025weakly} and mDeBERTa-v3-base-mnli-xnli \cite{ta2022transfer} by replacing their three-way classification heads with a binary classifier. This approach allowed us to leverage the semantic reasoning abilities these models had already acquired while adapting them specifically to our task. We do not report the zero-shot performance for these models, as the neural label cannot be directly mapped to contradiction or entailment. Moreover, we observed that the predicted label in the zero-shot setting was sensitive to the order of the input claims. For these reasons, we report only the results of the fine-tuned binary classifiers.

\paragraph{Large Language Models.} We experiment with the following multilingual LLMs for identifying contradictory visual claims: Phi-4-mini-instruct (Phi-4) \cite{abouelenin2025phi}, Mistral-7B-Instruct-v0.3 (Mistral) \cite{jiang2023mistral7b}, Llama-3.1-8B-Instruct (Llama3) \cite{dubey2024llama}, Gemma-7b (Gemma) \cite{team2024gemma}, and Qwen2.5-7B-Instruct (Qwen) \cite{yang2024qwen2_5}. We evaluated the models in zero-shot and fine-tuned settings. In the zero-shot scenario, the pre-trained models were instructed to predict class labels without any task-specific training. In both settings, we used two different prompts for monolingual and multilingual language configurations. For the multilingual setup, original languages of claim pairs are provided in the multilingual prompt to guide LLMs.

\subsection{Metrics}
Except for the zero-shot LLM setting, all other models were fine-tuned on the \textit{MultiCaption} train set and evaluated on \textit{MultiCaption} test set and \textit{COSMOS}. Zero-shot LLMs and fine-tuned model with different random seeds was evaluated five times, and we report the mean precision, recall, F1-score, and accuracy across these runs.

\section{Results}
\label{results}
\subsection{Baseline Performance}

\begin{table}[!t]
\centering
\scriptsize
\begin{tabular}{clcccc}
\hline
 & \textbf{Baseline} & \textbf{Precision} & \textbf{Recall} & \textbf{F1-Score} & \textbf{Accuracy}\\
\hline
\multicolumn{6}{c}{\textbf{COSMOS}}\\
\hline
\multirow{2}{*}{\makecell{Finetuned\\Transformer}} & XLM-R & $0.707$ & $0.425$ & $0.530$ & $0.625$\\
 & mDeBERTa & $0.712$ & $0.421$ & $0.528$ & $0.625$\\
 & mBERT & $0.653$ & $0.405$ & $0.500$ & $0.595$\\ \hline
\multirow{2}{*}{\makecell{Finetuned\\NLI}} & XLM-R & $0.844$ & $0.623$ & $0.717$ & $0.754$\\
 & mDeBERTa & $\mathbf{0.864}$ & $0.578$ & $0.693$ & $0.744$\\ \hline
\multirow{5}{*}{\makecell{Zero-shot\\LLM}} & Phi-4 & $0.502$ & $0.988$ & $0.665$ & $0.503$\\
 & Mistral & $0.789$ & $0.882$ & $0.833$ & $0.823$\\
 & Llama3 & $0.673$ & $0.805$ & $0.733$ & $0.706$\\
 & Gemma & $0.500$ & $\mathbf{1.000}$ & $0.667$ & $0.500$\\
 & Qwen & $0.706$ & $0.626$ & $0.663$ & $0.683$\\ \hline
\multirow{5}{*}{\makecell{Finetuned\\LLM}}  & Phi-4 & $0.696$ & $0.977$ & $0.814$ & $0.776$\\
 & Mistral & $0.774$ & $0.945$ & $\mathbf{0.851}$ & $\mathbf{0.834}$\\
 & Llama3 & $0.699$ & $0.963$ & $0.810$ & $0.774$\\
 & Gemma & $0.546$ & $0.992$ & $0.704$ & $0.582$\\
 & Qwen & $0.536$ & $0.994$ & $0.697$ & $0.567$\\ \hline
\multicolumn{6}{c}{\textbf{MultiCaption — Test (Monolingual)}}\\
\hline
 \multirow{2}{*}{\makecell{Finetuned\\Transformer}} & XLM-R & $\mathbf{0.894}$ & $0.729$ & $0.802$ & $0.824$\\
 & mDeBERTa & $0.879$ & $0.767$ & $0.819$ & $0.833$\\
 & mBERT & $0.884$ & $0.675$ & $0.766$ & $0.797$\\ \hline
\multirow{2}{*}{\makecell{Finetuned\\NLI}} & XLM-R & $0.795$ & $0.562$ & $0.659$ & $0.714$\\
 & mDeBERTa & $0.808$ & $0.654$ & $0.723$ & $0.754$\\ \hline
\multirow{5}{*}{\makecell{Zero-shot\\LLM}} & Phi-4 & $0.490$ & $0.990$ & $0.656$ & $0.489$\\
 & Mistral & $0.584$ & $0.954$ & $0.725$ & $0.644$\\
 & Llama3 & $0.515$ & $0.973$ & $0.673$ & $0.536$\\
 & Gemma & $0.491$ & $\mathbf{1.000}$ & $0.658$ & $0.491$\\
 & Qwen & $0.558$ & $0.782$ & $0.651$ & $0.589$\\ \hline
\multirow{5}{*}{\makecell{Finetuned\\LLM}}  & Phi-4 & $0.671$ & $0.965$ & $0.791$ & $0.751$\\
 & Mistral & $0.878$ & $0.949$ & $\mathbf{0.912}$ & $\mathbf{0.910}$\\
 & Llama3 & $0.796$ & $0.981$ & $0.878$ & $0.867$\\
 & Gemma & $0.724$ & $0.989$ & $0.835$ & $0.806$\\
 & Qwen & $0.593$ & $0.995$ & $0.743$ & $0.662$\\ \hline
\multicolumn{6}{c}{\textbf{MultiCaption — Test (Multilingual)}}\\
\hline
 \multirow{2}{*}{\makecell{Finetuned\\Transformer}} & XLM-R & $\mathbf{0.956}$ & $0.715$ & $0.816$ & $0.844$\\
 & mDeBERTa & $0.951$ & $0.767$ & $0.849$ & $0.866$\\
 & mBERT & $0.952$ & $0.747$ & $0.837$ & $0.857$\\ \hline
\multirow{2}{*}{\makecell{Finetuned\\NLI}} & XLM-R & $0.828$ & $0.518$ & $0.638$ & $0.710$\\
 & mDeBERTa & $0.837$ & $0.619$ & $0.712$ & $0.754$\\ \hline
\multirow{5}{*}{\makecell{Zero-shot\\LLM}} & Phi-4 & $0.489$ & $0.990$ & $0.655$ & $0.487$\\
 & Mistral & $0.528$ & $0.975$ & $0.685$ & $0.560$\\
 & Llama3 & $0.508$ & $0.984$ & $0.669$ & $0.524$\\
 & Gemma & $0.491$ & $\mathbf{1.000}$ & $0.658$ & $0.491$\\
 & Qwen & $0.522$ & $0.723$ & $0.606$ & $0.539$\\ \hline
\multirow{5}{*}{\makecell{Finetuned\\LLM}}  & Phi-4 & $0.533$ & $0.997$ & $0.695$ & $0.570$\\
 & Mistral & $0.866$ & $0.964$ & $0.912$ & $0.910$\\
 & Llama3 & $0.672$ & $0.992$ & $0.801$ & $0.758$\\
 & Gemma & $0.87$ & $0.97$ & $\mathbf{0.917}$ & $\mathbf{0.913}$\\
 & Qwen & $0.711$ & $0.985$ & $0.826$ & $0.796$\\
\hline
\end{tabular}
\caption{Baseline performance on COSMOS \& MultiCaption}
\label{tab:baseline_performance}
\end{table}

Table \ref{tab:baseline_performance} reports the performance of the baseline and fine-tuned models on the \textit{COSMOS} dataset and on the \textit{MultiCaption} test partition. The results show that fine-tuned NLI models outperform the fine-tuned transformer baselines on \textit{COSMOS}, suggesting that the NLI pretraining provides useful transferable knowledge for this dataset. However, on \textit{MultiCaption}, these NLI-based models underperform compared to trasnformer models suggesting that \textit{MultiCaption} poses challenges that extend beyond those captured in standard NLI tasks—even after fine-tuning. Moreover, none of the transformer-based classifiers exceed random-chance performance on \textit{COSMOS}. On \textit{MultiCaption}, however, the fine-tuned transformer model mDeBERTa achieves substantially higher F1-scores—both in monolingual and multilingual settings—surpassing many LLMs. This contrast highlights that these models adapt well to the data they are fine-tuned on, excelling when the task aligns well with their learned representations but struggling when it diverges.

A comparison of zero-shot LLM performance on \textit{MultiCaption} and \textit{COSMOS} reveals that, for some models, \textit{MultiCaption} is the more challenging dataset. For instance, Mistral achieves an F1-score of 0.83 on \textit{COSMOS} but drops to 0.72 on \textit{MultiCaption}, with Llama-3 showing a similar pattern. However, once fine-tuned, all LLMs show substantial performance gains, underscoring the effectiveness of task-specific fine-tuning.

Overall, the fine-tuned LLMs achieve the strongest performance on both datasets regardless of language configuration. Mistral, in particular, performs consistently well, reaching an F1-score of 0.851 on \textit{COSMOS} and 0.912 on \textit{MultiCaption}. Notably, its performance remains stable even when trained and evaluated on claims in their original languages. Other fine-tuned LLMs, such as Gemma and Qwen, show substantial gains from multilingual training, with improvements of nearly 8\% in F1-score. A similar trend is observed among multilingual fine-tuned transformer models, whose performance improves when trained and tested using the original, multilingual context. This underscores the value of multilingual datasets such as  \textit{MultiCaption}, which enable models to better handle linguistic diversity while supporting more efficient multilingual fact-checking pipelines without costly content translation.

\subsection{Performance Analysis}
\begin{figure}[t]
    \centering
    \begin{subfigure}[b]{0.6\linewidth}
        \centering
        \includegraphics[width=\linewidth]{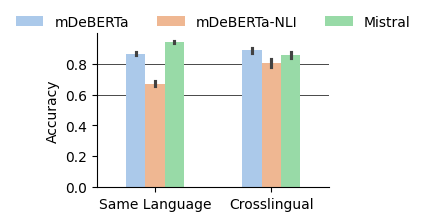}
        \caption{Same language vs. Crosslingual}
        \label{fig:accuracy_by_lan_pair}
    \end{subfigure}
    \hfill
    \begin{subfigure}[b]{0.7\linewidth}
        \centering
        \includegraphics[width=\linewidth]{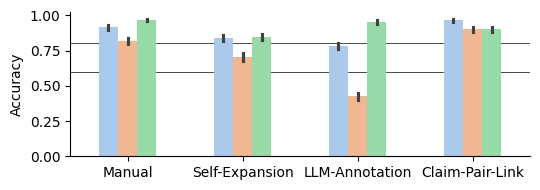}
        \caption{Label Strategy}
        \label{fig:accuracy_by_label_strategy}
    \end{subfigure}

    \begin{subfigure}[b]{1\linewidth}
        \centering
        \includegraphics[width=\linewidth]{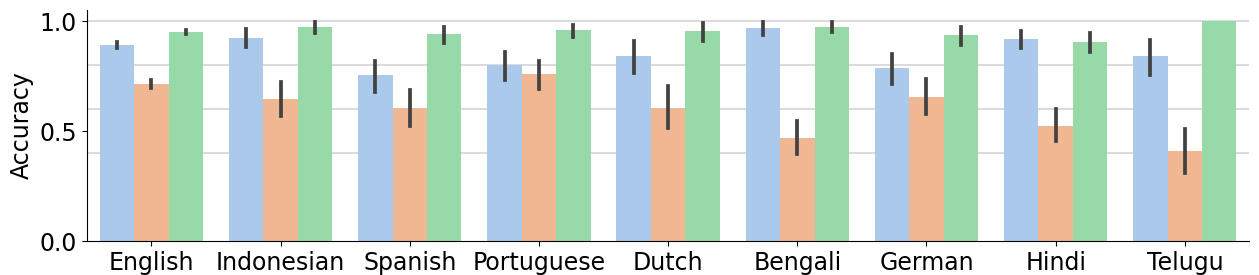}
        \caption{Language}
        \label{fig:accuracy_by_language}
    \end{subfigure}

    \caption{Accuracy of baselines across different settings: (a) same language vs. crosslingual, (b) labeling strategies, (c) language-specific accuracy.}
    \label{fig:combined_accuracy}
\end{figure}

We compare the accuracy of the best-performing baselines from each category—fine-tuned transformers, NLI models, and LLMs —under multilingual configuration across different settings of the \textit{MultiCaption} dataset. The results are shown in Figure \ref{fig:combined_accuracy}. The comparison reveals that mDeBERTa-NLI achieves lower accuracy when predicting pairs from the same language (Figure \ref{fig:accuracy_by_lan_pair}) as well on LLM annotated samples (Figure \ref{fig:accuracy_by_label_strategy}). This is likely because most contradictory pairs fall into these two categories, and NLI models often struggle to correctly identify contradiction. In contrast, the fine-tuned Mistral model maintains more consistent performance across both settings.
Figure \ref{fig:accuracy_by_language} presents the accuracy of the models by language on monolingual pairs for languages with at least 150 claims in the test set. Again, Mistral is consistent, being the top performer in all languages. In contrast, mDeBERTa-NLI exhibits substantial variability, with performance dropping sharply for some languages. These results also suggest that fine-tuning a transformer model from scratch can yield better task-specific adaptation than trying to leverage previously learned knowledge from NLI models for robust multilingual performance.

\section{Discussion and Conclusion}

This paper tackles the problem of identifying misinformation through contradictory visual claims, for which we first introduce \textit{MultiCaption}, the first multilingual dataset for this purpose. We employ multiple annotation strategies to label claim pairs about an image or video as contradicting or non-contradicting. The dataset contains 11,088 visual claim pairs written in 64 languages, making \textit{MultiCaption} a valuable resource for multilingual and crosslingual research.

Given the verified context of an image or video, \textit{MultiCaption} enables the detection of misinformation by identifying claims that contradict the original context. The inclusion of timestamps for each claim further supports applications such as analyzing the temporal and geographical diffusion of misinformation around the same piece of media. In this work, we focus exclusively on visual claim pairs and leave multimodal extensions for future research. Nevertheless, our baseline experiments demonstrate that textual information alone is sufficient in most cases to detect contradictory visual claims. Adding vision models could potentially increase detection capabilities, but they also come with additional computational cost. Our work shows that smaller language models, like mDeBERTa, can be a strong option for environments with limited resources.

We conduct extensive experiments to establish strong baselines using transformer models, natural language inference (NLI) models, and large language models (LLMs). Overall, our results show that \textit{MultiCaption} poses a greater challenge than standard NLI tasks: both zero-shot LLMs and fine-tuned NLI models struggle to achieve high performance. In contrast, fine-tuned transformer models and LLMs perform significantly better, underscoring the importance of task-specific training. Notably, multilingual models trained and evaluated on multilingual data achieve strong results, highlighting the dataset’s potential for building effective fact-checking pipelines without relying on machine translation. Our dataset and the fine-tuned models can easily be integrated in multilingual misinformation detection systems, allowing applications that involve claim verification and out-of-context detection. Experiments show that \textit{MultiCaption} is more challenging than a comparable existing dataset, incorporating the multilingual dimension and a larger volume of data. Future work will focus on extending both the dataset and methodology to multimodal settings.

Beyond the empirical results, the dataset construction methodology itself constitutes a significant contribution to the field of automated fact-checking. In particular, the use of claim–post links and the self-expansion of contradicting and non-contradicting pairs provide effective mechanisms for increasing dataset scale while introducing a multilingual dimension. Widely used disinformation datasets such as MultiClaim \cite{pikuliak2023multilingual}, MOCHEG \cite{yao2023end} and Factify \cite{mishra2022factify, suryavardan2023factify} could benefit from similar strategies, as they exploit latent links within already available data to expand coverage without requiring additional data collection, human annotation, or synthetic data generation.

\section{Limitations}
The limitations of this work are as follows:
\begin{itemize}
    \item A portion of the dataset is annotated using a Large Language Model (LLM). While possible annotation errors introduced by the LLM may propagate and affect the overall quality of the dataset, we tested the quality of the LLM annotations with a manually annotated dataset to mitigate this.
    \item Although the use of textual information is sufficient for the models to achieve high performance in contradiction detection, the approach can be extended to a multimodal setting. The current work is limited to text-only analysis.
    \item Detecting contradictions between two captions and determining which one constitutes misinformation requires access to the original or true context. Such context may not always be available, potentially necessitating manual intervention to identify the correct and incorrect captions.
    \item The construction of our dataset largely relies on the original data sources, inevitably resulting in an imbalanced distribution of samples across languages.  
\end{itemize}

\section{Acknowledgments}
Rafael Martins Frade and Rrubaa Panchendrarajan are funded by the European Union and UK Research and Innovation under Grant No. 101073351 as part of Marie Skłodowska-Curie Actions (MSCA Hybrid Intelligence to monitor, promote, and analyze transformations in good democracy practices). We acknowledge Queen Mary's Apocrita HPC facility, supported by QMUL Research-IT, for enabling our experiments \cite{king_2017_438045}.

\bibliography{references}

@article{multiclaimnet,
  title={MultiClaimNet: A Massively Multilingual Dataset of Fact-Checked Claim Clusters},
  author={Panchendrarajan, Rrubaa and M{\'\i}guez, Rub{\'e}n and Zubiaga, Arkaitz},
  journal={arXiv preprint arXiv:2503.22280},
  year={2025}
}

@inproceedings{devlin2019bert,
  title={Bert: Pre-training of deep bidirectional transformers for language understanding},
  author={Devlin, Jacob and Chang, Ming-Wei and Lee, Kenton and Toutanova, Kristina},
  booktitle={Proceedings of the 2019 conference of the North American chapter of the association for computational linguistics: human language technologies, volume 1 (long and short papers)},
  pages={4171--4186},
  year={2019}
}

@article{jiang2023mistral7b,
  title        = {Mistral 7B},
  author       = {Jiang, Albert Q. and Sablayrolles, Alexandre and Mensch, Arthur and Bamford, Chris and Chaplot, Devendra Singh and de las Casas, Diego and Bressand, Florian and Lengyel, Gianna and Lample, Guillaume and Saulnier, Lucile and others},
  journal      = {arXiv preprint arXiv:2310.06825},
  year         = {2023}
}

@inproceedings{pisarevskaya2025zero,
  title={Zero-shot and Few-shot Learning with Instruction-following LLMs for Claim Matching in Automated Fact-checking},
  author={Pisarevskaya, Dina and Zubiaga, Arkaitz},
  booktitle={Proceedings of the 31st International Conference on Computational Linguistics},
  pages={9721--9736},
  year={2025}
}

@article{chakraborty2023information,
  title={Information retrieval algorithms and neural ranking models to detect previously fact-checked information},
  author={Chakraborty, Tanmoy and La Gatta, Valerio and Moscato, Vincenzo and Sperl{\`\i}, Giancarlo},
  journal={Neurocomputing},
  volume={557},
  pages={126680},
  year={2023},
  publisher={Elsevier}
}

@manual{king_2017_438045,
  title        = {{Apocrita - High Performance Computing Cluster for 
                   Queen Mary University of London}},
  author       = {King, Thomas and
                  Butcher, Simon and
                  Zalewski, Lukasz},
  month        = mar,
  year         = 2017,
  doi          = {10.5281/zenodo.438045},
  url          = {https://doi.org/10.5281/zenodo.438045}
}

@article{gebru2021datasheets,
  title={Datasheets for datasets},
  author={Gebru, Timnit and Morgenstern, Jamie and Vecchione, Briana and Vaughan, Jennifer Wortman and Wallach, Hanna and Iii, Hal Daum{\'e} and Crawford, Kate},
  journal={Communications of the ACM},
  volume={64},
  number={12},
  pages={86--92},
  year={2021},
  publisher={ACM New York, NY, USA}
}

@misc{fair,
    title="The FAIR Data principles",
year = 2020,
    author="{FORCE11}",
howpublished="\url{https://force11.org/info/the-fair-data-principles/}"
}

@article{popordanoska2025clash,
  title={CLASH: A Benchmark for Cross-Modal Contradiction Detection},
  author={Popordanoska, Teodora and Li, Jiameng and Blaschko, Matthew B},
  journal={arXiv preprint arXiv:2511.19199},
  year={2025}
}

@misc{afp_example,
	author = {},
	title = {},
	howpublished = {\url{https://factcheck.afp.com/doc.afp.com.33LV9QC}},
	year = {2025},
	note = {[Accessed 29-11-2025]},
}

@article{senouci2025claim,
  title={Claim polarity analysis from conflicting sources},
  author={Senouci, Amel and Meziane, Hassina and Benbernou, Salima},
  journal={International Journal of Data Science and Analytics},
  volume={19},
  number={3},
  pages={435--451},
  year={2025},
  publisher={Springer}
}

@article{li2017contradiction,
  title={Contradiction detection with contradiction-specific word embedding},
  author={Li, Luyang and Qin, Bing and Liu, Ting},
  journal={Algorithms},
  volume={10},
  number={2},
  pages={59},
  year={2017},
  publisher={MDPI}
}

@inproceedings{li2024contradoc,
  title={Contradoc: understanding self-contradictions in documents with large language models},
  author={Li, Jierui and Raheja, Vipul and Kumar, Dhruv},
  booktitle={Proceedings of the 2024 Conference of the North American Chapter of the Association for Computational Linguistics: Human Language Technologies (Volume 1: Long Papers)},
  pages={6509--6523},
  year={2024}
}

@article{xu2024sparsecl,
  title={SparseCL: Sparse Contrastive Learning for Contradiction Retrieval},
  author={Xu, Haike and Lin, Zongyu and Sun, Yizhou and Chang, Kai-Wei and Indyk, Piotr},
  journal={arXiv preprint arXiv:2406.10746},
  year={2024}
}

@article{singh2023finding,
  title={Finding Already Debunked Narratives via Multistage Retrieval: Enabling Cross-Lingual, Cross-Dataset and Zero-Shot Learning},
  author={Singh, Iknoor and Scarton, Carolina and Song, Xingyi and Bontcheva, Kalina},
  journal={arXiv e-prints},
  pages={arXiv--2308},
  year={2023}
}

@inproceedings{nielsen2022mumin,
  title={Mumin: A large-scale multilingual multimodal fact-checked misinformation social network dataset},
  author={Nielsen, Dan S and McConville, Ryan},
  booktitle={Proceedings of the 45th International ACM SIGIR Conference on Research and Development in Information Retrieval},
  pages={3141--3153},
  year={2022}
}

@article{larraz2023semantic,
  title={Semantic similarity models for automated fact-checking: ClaimCheck as a claim matching tool},
  author={Larraz, Irene and M{\'\i}guez, Rub{\'e}n and Sallicati, Francesca},
  journal={Profesional de la informaci{\'o}n},
  volume={32},
  number={3},
  year={2023}
}

@article{panchendrarajan2024claim,
  title={Claim detection for automated fact-checking: A survey on monolingual, multilingual and cross-lingual research},
  author={Panchendrarajan, Rrubaa and Zubiaga, Arkaitz},
  journal={Natural Language Processing Journal},
  volume={7},
  pages={100066},
  year={2024},
  publisher={Elsevier}
}

@misc{yang2024qwen2_5,
  title        = {Qwen2.5 Technical Report},
  author       = {Yang, An and Yang, Baosong and Zhang, Beichen and Hui, Binyuan and Zheng, Bo and Yu, Bowen and Li, Chengyuan and Liu, Dayiheng and Huang, Fei and Wei, Haoran and Lin, Huan and Yang, Jian and Tu, Jianhong and Zhang, Jianwei and Yang, Jianxin and Yang, Jiaxi and Zhou, Jingren and Lin, Junyang and Dang, Kai and Lu, Keming and Bao, Keqin and Yang, Kexin and Yu, Le and Li, Mei and Xue, Mingfeng and Zhang, Pei and Zhu, Qin and Men, Rui and Lin, Runji and Li, Tianhao and Tang, Tianyi and Xia, Tingyu and Ren, Xingzhang and Ren, Xuancheng and Fan, Yang and Su, Yang and Zhang, Yichang and Wan, Yu and Liu, Yuqiong and Cui, Zeyu and Zhang, Zhenru and Qiu, Zihan and others},
  howpublished = {arXiv preprint arXiv:2412.15115},
  year         = {2024}
}

@article{team2024gemma,
  title={Gemma: Open models based on gemini research and technology},
  author={Team, Gemma and Mesnard, Thomas and Hardin, Cassidy and Dadashi, Robert and Bhupatiraju, Surya and Pathak, Shreya and Sifre, Laurent and Rivi{\`e}re, Morgane and Kale, Mihir Sanjay and Love, Juliette and others},
  journal={arXiv preprint arXiv:2403.08295},
  year={2024}
}

@article{dubey2024llama,
  title={The llama 3 herd of models},
  author={Dubey, Abhimanyu and Jauhri, Abhinav and Pandey, Abhinav and Kadian, Abhishek and Al-Dahle, Ahmad and Letman, Aiesha and Mathur, Akhil and Schelten, Alan and Yang, Amy and Fan, Angela and others},
  journal={arXiv e-prints},
  pages={arXiv--2407},
  year={2024}
}

@article{abouelenin2025phi,
  title={Phi-4-mini technical report: Compact yet powerful multimodal language models via mixture-of-loras},
  author={Abouelenin, Abdelrahman and Ashfaq, Atabak and Atkinson, Adam and Awadalla, Hany and Bach, Nguyen and Bao, Jianmin and Benhaim, Alon and Cai, Martin and Chaudhary, Vishrav and Chen, Congcong and others},
  journal={arXiv preprint arXiv:2503.01743},
  year={2025}
}

@inproceedings{ta2022transfer,
  title={Transfer Learning from Multilingual DeBERTa for Sexism Identification.},
  author={Ta, Hoang Thang and Rahman, Abu Bakar Siddiqur and Najjar, Lotfollah and Gelbukh, Alexander F},
  booktitle={IberLEF@ SEPLN},
  year={2022}
}

@article{alotaibi2025weakly,
  title={Weakly Supervised Deep Learning for Arabic Tweet Sentiment Analysis on Education Reforms: Leveraging Pre-trained Models and LLMs with Snorkel},
  author={Alotaibi, Alanoud and Nadeem, Farrukh and Hamdy, Mohamed},
  journal={IEEE Access},
  year={2025},
  publisher={IEEE}
}

@inproceedings{aneja2023cosmos,
  title={COSMOS: catching out-of-context image misuse using self-supervised learning},
  author={Aneja, Shivangi and Bregler, Chris and Nie{\ss}ner, Matthias},
  booktitle={Proceedings of the AAAI conference on artificial intelligence},
  volume={37},
  number={12},
  pages={14084--14092},
  year={2023}
}

@inproceedings{pikuliak2023multilingual,
  title={Multilingual Previously Fact-Checked Claim Retrieval},
  author={Pikuliak, Mat{\'u}{\v{s}} and Srba, Ivan and Moro, Robert and Hromadka, Timo and Smole{\v{n}}, Timotej and Meli{\v{s}}ek, Martin and Vykopal, Ivan and Simko, Jakub and Podrou{\v{z}}ek, Juraj and Bielikov{\'a}, M{\'a}ria},
  booktitle={Proceedings of the 2023 Conference on Empirical Methods in Natural Language Processing},
  pages={16477--16500},
  year={2023}
}

@article{sepulveda2023detecting,
  title={Detecting misleading headlines through the automatic recognition of contradiction in spanish},
  author={Sep{\'u}lveda-Torres, Robiert and Bonet-Jover, Alba and Saquete, Estela},
  journal={IEEE Access},
  volume={11},
  pages={72007--72026},
  year={2023},
  publisher={IEEE}
}

@inproceedings{reimers-2019-sentence-bert,
  title = "Sentence-BERT: Sentence Embeddings using Siamese BERT-Networks",
  author = "Reimers, Nils and Gurevych, Iryna",
  booktitle = "Proceedings of the 2019 Conference on Empirical Methods in Natural Language Processing",
  month = "11",
  year = "2019",
  publisher = "Association for Computational Linguistics",
  url = "https://arxiv.org/abs/1908.10084",
}

@inproceedings{conneau2020unsupervised,
  title={Unsupervised cross-lingual representation learning at scale},
  author={Conneau, Alexis and Khandelwal, Kartikay and Goyal, Naman and Chaudhary, Vishrav and Wenzek, Guillaume and Guzm{\'a}n, Francisco and Grave, Edouard and Ott, Myle and Zettlemoyer, Luke and Stoyanov, Veselin},
  booktitle={Proceedings of the 58th annual meeting of the association for computational linguistics},
  pages={8440--8451},
  year={2020}
}

@article{he2021debertav3,
  title={Debertav3: Improving deberta using electra-style pre-training with gradient-disentangled embedding sharing},
  author={He, Pengcheng and Gao, Jianfeng and Chen, Weizhu},
  journal={arXiv preprint arXiv:2111.09543},
  year={2021}
}

@article{papadopoulos2024verite,
  title={Verite: a robust benchmark for multimodal misinformation detection accounting for unimodal bias},
  author={Papadopoulos, Stefanos-Iordanis and Koutlis, Christos and Papadopoulos, Symeon and Petrantonakis, Panagiotis C},
  journal={International Journal of Multimedia Information Retrieval},
  volume={13},
  number={1},
  pages={4},
  year={2024},
  publisher={Springer}
}

@inproceedings{abdelnabi2022open,
  title={Open-domain, content-based, multi-modal fact-checking of out-of-context images via online resources},
  author={Abdelnabi, Sahar and Hasan, Rakibul and Fritz, Mario},
  booktitle={Proceedings of the IEEE/CVF conference on computer vision and pattern recognition},
  pages={14940--14949},
  year={2022}
}

@inproceedings{nguyen2023multi,
  title={Multi-Models from Computer Vision to Natural Language Processing for Cheapfakes Detection},
  author={Nguyen, Thanh--Son and Tran, Minh--Triet},
  booktitle={2023 IEEE International Conference on Multimedia and Expo Workshops (ICMEW)},
  pages={93--98},
  year={2023},
  organization={IEEE}
}

@inproceedings{la2022combination,
  title={A Combination of Visual-Semantic Reasoning and Text Entailment-based Boosting Algorithm for Cheapfake Detection},
  author={La, Tuan-Vinh and Dao, Minh-Son and Tran, Quang-Tien and Tran, Thanh-Phuc and Tran, Anh-Duy and Dang-Nguyen, Duc-Tien},
  booktitle={Proceedings of the 30th ACM International Conference on Multimedia},
  pages={7140--7144},
  year={2022}
}

@inproceedings{tran2022textual,
  title={A Textual-Visual-Entailment-based Unsupervised Algorithm for Cheapfake Detection},
  author={Tran, Quang-Tien and Tran, Thanh-Phuc and Dao, Minh-Son and La, Tuan-Vinh and Tran, Anh-Duy and Dang Nguyen, Duc Tien},
  booktitle={Proceedings of the 30th ACM International Conference on Multimedia},
  pages={7145--7149},
  year={2022}
}

@article{la2022leverage,
  title={Leverage Boosting and Transformer on Text-Image Matching for Cheap Fakes Detection},
  author={La, Tuan-Vinh and Dao, Minh-Son and Le, Duy-Dong and Thai, Kim-Phung and Nguyen, Quoc-Hung and Phan-Thi, Thuy-Kieu},
  journal={Algorithms},
  volume={15},
  number={11},
  pages={423},
  year={2022},
  publisher={MDPI}
}

@inproceedings{nguyen2022grit,
  title={Grit: Faster and better image captioning transformer using dual visual features},
  author={Nguyen, Van-Quang and Suganuma, Masanori and Okatani, Takayuki},
  booktitle={European Conference on Computer Vision},
  pages={167--184},
  year={2022},
  organization={Springer}
}

@article{gubelmann2024capturing,
  title={Capturing the varieties of natural language inference: A systematic survey of existing datasets and two novel benchmarks},
  author={Gubelmann, Reto and Katis, Ioannis and Niklaus, Christina and Handschuh, Siegfried},
  journal={Journal of Logic, Language and Information},
  volume={33},
  number={1},
  pages={21--48},
  year={2024},
  publisher={Springer}
}

@inproceedings{ahmad2025climatecheck,
  title={The ClimateCheck shared task: Scientific fact-checking of social media claims about climate change},
  author={Ahmad, Raia Abu and Usmanova, Aida and Rehm, Georg},
  booktitle={Proceedings of the Fifth Workshop on Scholarly Document Processing (SDP 2025)},
  pages={263--275},
  year={2025}
}

@article{dmonte2024claim,
  title={Claim verification in the age of large language models: A survey},
  author={Dmonte, Alphaeus and Oruche, Roland and Zampieri, Marcos and Calyam, Prasad and Augenstein, Isabelle},
  journal={arXiv preprint arXiv:2408.14317},
  year={2024}
}

@article{thorne2018fever,
  title={FEVER: a large-scale dataset for fact extraction and VERification},
  author={Thorne, James and Vlachos, Andreas and Christodoulopoulos, Christos and Mittal, Arpit},
  journal={arXiv preprint arXiv:1803.05355},
  year={2018}
}

@article{aly2021feverous,
  title={Feverous: Fact extraction and verification over unstructured and structured information},
  author={Aly, Rami and Guo, Zhijiang and Schlichtkrull, Michael and Thorne, James and Vlachos, Andreas and Christodoulopoulos, Christos and Cocarascu, Oana and Mittal, Arpit},
  journal={arXiv preprint arXiv:2106.05707},
  year={2021}
}

@article{akhtar2023multimodal,
  title={Multimodal automated fact-checking: A survey},
  author={Akhtar, Mubashara and Schlichtkrull, Michael and Guo, Zhijiang and Cocarascu, Oana and Simperl, Elena and Vlachos, Andreas},
  journal={arXiv preprint arXiv:2305.13507},
  year={2023}
}

@inproceedings{elsayed2019overview,
  title={Overview of the CLEF-2019 CheckThat! Lab: automatic identification and verification of claims},
  author={Elsayed, Tamer and Nakov, Preslav and Barr{\'o}n-Cedeno, Alberto and Hasanain, Maram and Suwaileh, Reem and Da San Martino, Giovanni and Atanasova, Pepa},
  booktitle={International conference of the cross-language evaluation forum for European languages},
  pages={301--321},
  year={2019},
  organization={Springer}
}

@article{grootendorst2022bertopic,
  title={BERTopic: Neural topic modeling with a class-based TF-IDF procedure},
  author={Grootendorst, Maarten},
  journal={arXiv preprint arXiv:2203.05794},
  year={2022}
}

@inproceedings{yao2023end,
  title={End-to-end multimodal fact-checking and explanation generation: A challenging dataset and models},
  author={Yao, Barry Menglong and Shah, Aditya and Sun, Lichao and Cho, Jin-Hee and Huang, Lifu},
  booktitle={Proceedings of the 46th International ACM SIGIR Conference on Research and Development in Information Retrieval},
  pages={2733--2743},
  year={2023}
}

@inproceedings{mishra2022factify,
  title={FACTIFY: A Multi-Modal Fact Verification Dataset.},
  author={Mishra, Shreyash and Suryavardan, S and Bhaskar, Amrit and Chopra, Parul and Reganti, Aishwarya N and Patwa, Parth and Das, Amitava and Chakraborty, Tanmoy and Sheth, Amit P and Ekbal, Asif and others},
  booktitle={DE-FACTIFY@ AAAI},
  year={2022}
}

@article{suryavardan2023factify,
  title={Factify 2: A multimodal fake news and satire news dataset},
  author={Suryavardan, S and Mishra, Shreyash and Patwa, Parth and Chakraborty, Megha and Rani, Anku and Reganti, Aishwarya and Chadha, Aman and Das, Amitava and Sheth, Amit and Chinnakotla, Manoj and others},
  journal={arXiv preprint arXiv:2304.03897},
  year={2023}
}

@article{luo2021newsclippings,
  title={Newsclippings: Automatic generation of out-of-context multimodal media},
  author={Luo, Grace and Darrell, Trevor and Rohrbach, Anna},
  journal={arXiv preprint arXiv:2104.05893},
  year={2021}
}

@inproceedings{papadopoulos2025similarity,
  title={Similarity over Factuality: Are we making progress on multimodal out-of-context misinformation detection?},
  author={Papadopoulos, Stefanos-Iordanis and Koutlis, Christos and Papadopoulos, Symeon and Petrantonakis, Panagiotis C},
  booktitle={2025 IEEE/CVF Winter Conference on Applications of Computer Vision (WACV)},
  pages={5041--5050},
  year={2025},
  organization={IEEE}
}

\subsection{Paper Checklist}

\begin{enumerate}

\item For most authors...
\begin{enumerate}
    \item  Would answering this research question advance science without violating social contracts, such as violating privacy norms, perpetuating unfair profiling, exacerbating the socio-economic divide, or implying disrespect to societies or cultures?
    \answerYes{Yes. The research advances automated fact-checking and misinformation detection without violating privacy norms or social contracts. The dataset is constructed from publicly available fact-checking articles and social media claims, does not include personal or sensitive private data, and aims to mitigate societal harm caused by disinformation.}
  \item Do your main claims in the abstract and introduction accurately reflect the paper's contributions and scope?
    \answerYes{Yes. The abstract and introduction accurately describe the creation of a multilingual dataset for contradictory visual claims, the labeling strategies employed, and the experimental evaluation of multiple model families, which align with the paper’s actual contributions.}
   \item Do you clarify how the proposed methodological approach is appropriate for the claims made? 
    \answerYes{Yes. The paper clearly motivates contradiction detection as a suitable formulation for identifying visual misinformation and justifies the use of data sources, labeling strategies, and multilingual modeling.}
   \item Do you clarify what are possible artifacts in the data used, given population-specific distributions?
    \answerYes{Yes. Refer to \textit{MultiCaption Dataset Construction}.}
  \item Did you describe the limitations of your work?
    \answerYes{Yes. Refer to \textit{Limitations}.}
  \item Did you discuss any potential negative societal impacts of your work?
    \answerYes{Yes. Refer to \textit{Discussion and Conclusion} and \textit{Limitations}.}
      \item Did you discuss any potential misuse of your work?
    \answerYes{Yes. Refer to \textit{Limitations}.}
    \item Did you describe steps taken to prevent or mitigate potential negative outcomes of the research, such as data and model documentation, data anonymization, responsible release, access control, and the reproducibility of findings?
    \answerYes{Yes. Refer to \textit{Baseline Implementation}. Source code will be released for reproducibility, and the dataset will be released with restricted access only for research purpose}
  \item Have you read the ethics review guidelines and ensured that your paper conforms to them?
    \answerYes{Yes. The paper conforms to ICWSM ethics guidelines.}
\end{enumerate}

\item Additionally, if your study involves hypotheses testing...
\begin{enumerate}
  \item Did you clearly state the assumptions underlying all theoretical results?
    \answerNA{NA. The study is empirical; no formal hypotheses are tested.}
  \item Have you provided justifications for all theoretical results?
    \answerNA{NA.}
  \item Did you discuss competing hypotheses or theories that might challenge or complement your theoretical results?
    \answerNA{NA.}
  \item Have you considered alternative mechanisms or explanations that might account for the same outcomes observed in your study?
    \answerYes{Yes. Refer to \textit{Results and Discussion and Conclusion.}}
  \item Did you address potential biases or limitations in your theoretical framework?
    \answerYes{Yes. Refer to \textit{Limitations.}}
  \item Have you related your theoretical results to the existing literature in social science?
    \answerYes{Yes. Refer to \textit{Related Work}.}
  \item Did you discuss the implications of your theoretical results for policy, practice, or further research in the social science domain?
    \answerYes{Yes. Refer to \textit{Discussion and Conclusion}.}
\end{enumerate}

\item Additionally, if you are including theoretical proofs...
\begin{enumerate}
  \item Did you state the full set of assumptions of all theoretical results?
    \answerNA{NA.}
	\item Did you include complete proofs of all theoretical results?
    \answerNA{NA.}
\end{enumerate}

\item Additionally, if you ran machine learning experiments...
\begin{enumerate}
  \item Did you include the code, data, and instructions needed to reproduce the main experimental results (either in the supplemental material or as a URL)?
    \answerYes{Yes. Source code and dataset are provided as supplementary materials and will be published upon acceptance.}
  \item Did you specify all the training details (e.g., data splits, hyperparameters, how they were chosen)?
    \answerYes{Yes. Refer to \textit{Experiment Setup and Baseline Implementation in Appendix}.}
     \item Did you report error bars (e.g., with respect to the random seed after running experiments multiple times)?
    \answerYes{Yes. Refer to \textit{Metrics}.}
	\item Did you include the total amount of compute and the type of resources used (e.g., type of GPUs, internal cluster, or cloud provider)?
    \answerYes{Yes. Refer to \textit{Baseline Implementation in Appendix}.}
     \item Do you justify how the proposed evaluation is sufficient and appropriate to the claims made? 
    \answerYes{Yes. Refer to \textit{Experiment Setup and Results}.}
     \item Do you discuss what is ``the cost`` of misclassification and fault (in)tolerance?
    \answerYes{Yes. Refer to \textit{Discussion and Conclusion}.}
  
\end{enumerate}

\item Additionally, if you are using existing assets (e.g., code, data, models) or curating/releasing new assets, \textbf{without compromising anonymity}...
\begin{enumerate}
  \item If your work uses existing assets, did you cite the creators?
    \answerYes{Yes. Refer to \textit{MultiCaption Dataset Construction}.}
  \item Did you mention the license of the assets?
    \answerYes{Yes. Refer to \textit{Artifact Availability} in Appendix.} 
  \item Did you include any new assets in the supplemental material or as a URL?
    \answerYes{Yes. Refer to \textit{Artifact Availability} in Appendix.} 
  \item Did you discuss whether and how consent was obtained from people whose data you're using/curating?
    \answerYes{Yes. Refer to \textit{Artifact Availability} in Appendix.} 
  \item Did you discuss whether the data you are using/curating contains personally identifiable information or offensive content?
    \answerYes{Yes. Refer to \textit{Artifact Availability} in Appendix.} 
\item If you are curating or releasing new datasets, did you discuss how you intend to make your datasets FAIR (see \citet{fair})?
\answerYes{Yes. Refer to \textit{Artifact Availability} in Appendix.} 
\item If you are curating or releasing new datasets, did you create a Datasheet for the Dataset (see \citet{gebru2021datasheets})? 
\answerNo{No. A formal datasheet is not included; dataset documentation is provided instead. Refer to \textit{Artifact Availability} in Appendix.}
\end{enumerate}

\item Additionally, if you used crowdsourcing or conducted research with human subjects, \textbf{without compromising anonymity}...
\begin{enumerate}
  \item Did you include the full text of instructions given to participants and screenshots?
    \answerNA{NA.}
  \item Did you describe any potential participant risks, with mentions of Institutional Review Board (IRB) approvals?
    \answerNA{NA.}
  \item Did you include the estimated hourly wage paid to participants and the total amount spent on participant compensation?
    \answerNA{NA.}
   \item Did you discuss how data is stored, shared, and deidentified?
   \answerNA{NA.}
\end{enumerate}

\end{enumerate}

\appendix
\section{Similarity distribution of non-contradicting claims created via claim-post links}\label{app:sim_dis_cp_links}
\begin{figure}[h]
    \centering
    \includegraphics[width=0.7\linewidth]{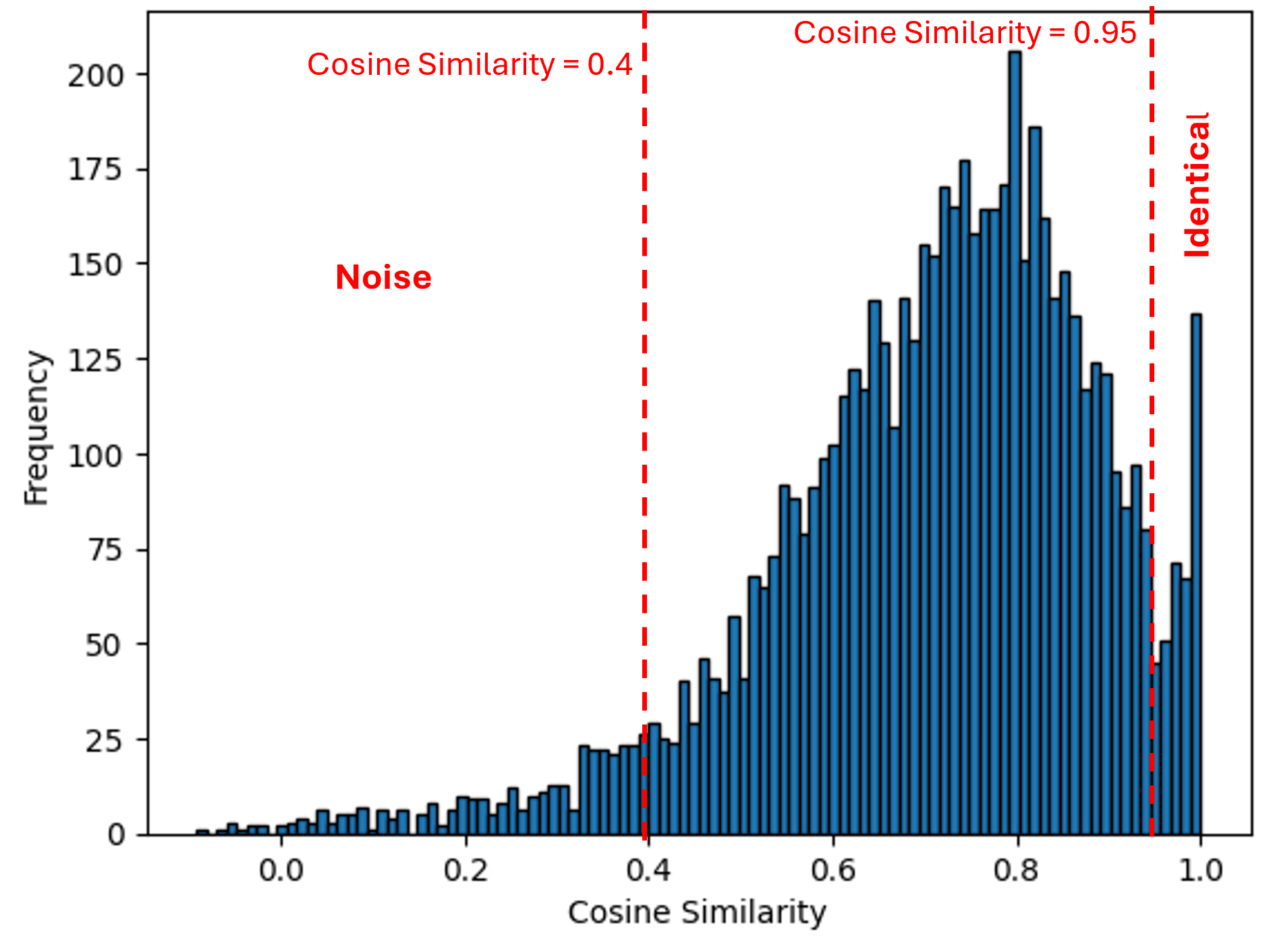}
    \caption{Similarity distribution of non-contradicting pairs generated via claim-post links}
    \label{fig:sim_dis_cp_links}
\end{figure}

Figure \ref{fig:sim_dis_cp_links} shows the similarity distribution of non-contradicting claim-pairs created via claim-post links and the discarded regions (noise and identical).

\section{Negation Terms for Filtering Fact-Checked Article Titles}


\begin{table}[h!]
\centering
\scriptsize
\caption{List of negation terms used to filter out titles with direct reference or denial of the claim}\label{tab:negation_terms}
\begin{tabular}{p{0.95\linewidth}}
\toprule
\textbf{Negation Terms} \\
\midrule
fake, ?, mislead, false, wrong, manipulat, edited, altered, not, no, photoshop, isn’t, purport, fabricated, forged, mistaken, claim that, doesn't, doesn’t, wasn't, watsn’t, morphed, doctored, misuse, unrelated, rumor, nothing, generated, misrepresent, misinterpret, didn’t, shared as, fact check:, fraudulent, deceive, misinform, misguide, invalid, inaccurate, untrue, erroneous, tamper, distort, modified, portrayed as, falsified, concoct, bogus, phony, hoax, associated with, aren’t, isn't, neither, miscaption, context, since, fictional, montage, ,not, actually, incorrect, old, retouched, reject, as if, doesnt, didnt, apocryphal, didnt, doesnt, yes, predate, before, previous, prior, another, none, far from, in reports about, satirical, attributed, was taken in, not, no, attention, never, claiming, staged, claim, joke, in fact, is from, are from, dates from, date from, but \\
\bottomrule
\end{tabular}
\end{table}


\section{LLM paraphrase generation of contradicting pairs}
\begin{prompt}[H]
\begin{scriptsize}
\begin{verbatim}
### Instruction:
You are given a list of image caption pairs. For each 
pair, generate a paraphrase of the {prefix_param} caption 
only. The paraphrase should keep the same meaning, but be 
as different as possible.
\end{verbatim}
  \caption{Paraphrase generation}
  \label{prompt-paraphrase}
\end{scriptsize}
\end{prompt}\label{sec:appendix_data_expansion}
Prompt \ref{prompt-paraphrase} presents the instruction used to generate the paraphrases using gpt-5-mini-2025-08-07. "prefix\_param" was set to either the claim or title to generate two new pairs of contradicting samples.

\section{LLM Annotation of Contradicting Pairs}\label{appendix:llm_annotation}

\begin{prompt}[H]
\begin{scriptsize}
\begin{verbatim}
### Instruction:
You are given two statements: 

Claim – A statement describing the content of an image or
video. Title – The title of a fact-checking article 
referring to that same image or video. 

Your task is to assign two binary labels: 
{"contradict": <True/False>, "denial": <True/False>} 

Contradict:
Mark True if the claim and title cannot both be true for
the same image or video — that is, they describe 
opposing or mutually exclusive facts (different events, 
people, locations, or circumstances). Mark False if they 
could both be true, or if the title simply adds 
compatible information. Contradict=True only when both 
cannot be true simultaneously, not when one is just less 
specific or a partial correction. 

Examples:
True → “Photo shows protest in India” vs “Photo shows 
protest in Bangladesh.”
True → “Taken in 2021” vs “Taken in 2019.”
False → “Photo shows protest in Delhi” vs “Photo shows 
Indian protest in Delhi.” (Compatible)

Denial:
Mark True if the title contains explicit refutation or 
debunking language, 
even if it also adds context. Typical denial cues 
include: - Direct negations: “does not,” “did not,” 
“is not,” “no, this is not...”
- Verdict terms: “false,” “fake,” “misleading,” 
“fabricated,” “debunked.”
- Authority denials: “officials denied,” “government 
refuted,” “police said it’s fake.”
- Refutational phrasing: “other photos show...,” “not 
what it claims to be.”
- Direct references: “original photo shows...”, “real 
video is ....”

Mark False if none of these cues appear. Denial=True 
only if explicit refutation words are present, not 
merely implied by a differing fact. 

Examples:
True → “Other photos from the event show them shaking 
hands.”
True → “Claim that this shows COVID-19 detention is 
false.”
False → “This video shows a 2017 airshow in Italy.”

Claim:  
Title: 
\end{verbatim}
  \caption{Contradiction and denial labeling prompt}
  \label{prompt-v1}
\end{scriptsize}
\end{prompt}

Prompt \ref{prompt-v1} presents annotation instructions supplied to ChatGPT (snapshot \textit{gpt-5-2025-08-07}) for assigning \textit{contradict} and \textit{denial} labels to candidate claim-title pairs. Table \ref{tab:llm_annotation_distribution} presents the distribution of \textit{contradict} and \textit{denial} labels assigned by the LLM. Among the 8197 pairs, approximately 43\% (\textit{denial}=True) were discarded because the title explicitly denied the claim, and 31\% were discarded as the title merely rephrased the claim (\textit{contradict}=False, \textit{denial}=False).

\begin{table}[H] 
\centering
\scriptsize
\caption{Distribution of contradict and denial combinations}\label{tab:llm_annotation_distribution}
\begin{tabular}{lllp{4cm}}
\hline
\rotatebox{90}{\textbf{Contradict}} &
\rotatebox{90}{\textbf{Denial}} &
\textbf{\# Pairs} &
\textbf{Majority Cases} \\ \hline
True & True & 3227 (39\%) & Title denies the fact-checked claim \\
\textbf{True} & \textbf{False} & \textbf{2072} (25\%) & \textbf{Title contains only the true claim} \\
False & True & 355 (4\%) & Title is a rephrase of fact-checked claim with denying phrases \\
False & False & 2543 (31\%) & Title is a rephrase of fact-checked claim \\ \hline
\end{tabular}
\end{table}

\section{Contradiction Detection}

\subsection{LLM Prompts}
Prompts \ref{prompt-cd-mon} and \ref{prompt-cd-mul} provide the instructions used for the LLMs to predict contradiction labels in monolingual and multilingual settings, respectively.

\begin{prompt}[t]
\begin{scriptsize}
\begin{verbatim}
You are an expert fact-checking assistant. 
Your task is to decide if two claims about the same image 
or video contradict each other.

Definition of Contradict:
Answer 'Yes' if the two claims cannot both be true for 
the same image or video that is, they describe opposing 
or mutually exclusive facts (different events, people, 
locations, or circumstances). Answer 'No' if they could 
both be true, or if one simply adds compatible 
information. Two claims are contradicting only when both 
cannot be true simultaneously, not when one is just less 
specific or a partial correction.
Claim 1: {claim1}
Claim 2: {claim2}

Question: Do Claim 1 and Claim 2 contradict each other 
according to the definition above? Answer with only one 
word: Yes or No
Answer: 
    
\end{verbatim}
  \caption{Contradiction detection - Monolingual}
  \label{prompt-cd-mon}
\end{scriptsize}
\end{prompt}

\begin{prompt}[t]
\begin{scriptsize}
\begin{verbatim}
You are an expert multilingual fact-checking assistant. 
Your task is to decide if two claims about the same image
or video contradict each other. The two claims below may 
be written in different languages, but always produce the 
final answer **in English only**.

Definition of Contradict:
Answer 'Yes' if the two claims cannot both be true for 
the same image or video that is, they describe opposing 
or mutually exclusive facts (different events, people, 
locations, or circumstances). Answer 'No' if they could 
both be true, or if one simply adds compatible 
information. Two claims are contradicting only when both 
cannot be true simultaneously, not when one is just less 
specific or a partial correction.
Claim 1 written in {language1}: {claim1}
Claim 2 written in {language2}: {claim2}

Question: Do Claim 1 and Claim 2 contradict each other 
according to the definition above? Answer with only one 
word in English: Yes or No
Answer: 
    
\end{verbatim}
  \caption{Contradiction detection - Multilingual}
  \label{prompt-cd-mul}
\end{scriptsize}
\end{prompt}

\subsection{Baseline Implementation}\label{appendix:baselines}

\paragraph{\textbf{Finetuning Transformers}}

Transformer models were trained with a learning rate of 2e-5, batch size 16, for 5 epochs, with 10\% of training data used for validation. The model with best validation F1-Score among the 5 epochs was saved. This process was performed 5 times for each model and the average results were the ones reported. Our experiments utilized the following implementations: XLM-Roberta-large (XLM-R) \footnote{https://huggingface.co/xlm-roberta-large}, Multilingual Deberta-V3 (mDeBERTa) \footnote{https://huggingface.co/microsoft/mdeberta-v3-large}, Multilingual BERT (mBERT) \footnote{https://huggingface.co/google-bert/bert-base-multilingual-cased}.

\begin{table}[t]
\caption{Hyperparameters used for LLM finetuning.}
\scriptsize
\centering
\label{tab:hyperparameters_llm}
\begin{tabular}{ll}
\hline
\textbf{Parameter} & \textbf{Value} \\ \hline

\multicolumn{2}{l}{\textit{LoRA Configuration}} \\

LoRA rank ($r$) & 64 \\
LoRA alpha & 16 \\
LoRA dropout & 0.1 \\
Target modules & all-linear \\
\multicolumn{2}{l}{\textit{Training Arguments}} \\
Epochs & 3 \\
Batch size (per device) & 1 \\
Gradient accumulation steps & 4 \\
Optimizer & paged\_adamw\_8bit \\
Learning rate & 2e--5 \\
Learning rate scheduler & constant \\
Weight decay & 0.001 \\
Warmup ratio & 0.03 \\
Max gradient norm & 0.3 \\ \hline
\end{tabular}
\end{table}

\paragraph{\textbf{Finetuning NLI Models}}
To leverage the existing knowledge of the models previously trained for NLI, we set a small learning (1e-6). This allowed models to retain prior classification abilities, while still adapting to the binary contradiction/non-contradiction task. The training and evaluation process was the same for the other transformer models. We conducted our experiments using these pretrained NLI models: XLM-RoBERTa-large-xnli (XLM-RoBERTa-NLI) \footnote{https://huggingface.co/joeddav/xlm-roberta-large-xnli} and mDeBERTa-v3-base-mnli-xnli (mDeBERTa-NLI) \footnote{https://huggingface.co/MoritzLaurer/mDeBERTa-v3-base-mnli-xnli} 

\paragraph{\textbf{Zero-shot LLMs}}\label{sec:appendix_llm_zeroshot}
We conduct all Large language model (LLM)-based evaluations on a single GPU with 8 cores, each equipped with 11 GB of memory. Prompts \ref{prompt-cd-mon} and \ref{prompt-cd-mul} specify the templates used for the monolingual and multilingual evaluation settings, respectively. The LLMs are instructed to generate a single token (“Yes” or “No”) with a temperature of 0.1. The following Hugging Face–hosted models were used in our experiments: microsoft/Phi-4-mini-instruct (Phi-4)\footnote{https://huggingface.co/microsoft/Phi-4-mini-instruct}, mistralai/Mistral-7B-Instruct-v0.3 (Mistral)\footnote{https://huggingface.co/mistralai/Mistral-7B-Instruct-v0.3}, meta-llama/Llama-3.1-8B-Instruct (Llama3) \footnote{https://huggingface.co/meta-llama/Llama-3.1-8B-Instruct}, google/gemma-7b (Gemma)\footnote{https://huggingface.co/google/gemma-7b}, Qwen/Qwen2.5-7B-Instruct (Qwen\footnote{https://huggingface.co/Qwen/Qwen2.5-7B-Instruct}.

\paragraph{\textbf{Finetuning LLMs}}
We fine-tuned the LLMs described earlier using the same prompts applied during testing. To improve memory efficiency, we employed LoRA (Low-Rank Adaptation), a parameter-efficient fine-tuning (PEFT) method, in combination with 4-bit quantization. Table \ref{tab:hyperparameters_llm} summarizes the hyperparameters used during fine-tuning.

\section{Artifact Availability}
Following research artifacts are released for reproducibility:

\begin{itemize}
    \item The \textit{MultiCaption} dataset\footnote{Dataset is available at \url{https://doi.org/10.5281/zenodo.18230659}}, including the finalized train and test splits used in all experiments.
    \item Source code\footnote{Source code is available at \url{https://github.com/rfrade/multicaption}} for data preprocessing, model training, and evaluation.
\end{itemize}

The source datasets \textit{MultiClaim} and \textit{MultiClaimNet} are available under restricted access for research purposes only, subject to their original licensing terms. The \textit{MultiCaption} is released under the same restricted-access conditions, permitting use exclusively for non-commercial research.

No original social media content is redistributed. All claims and metadata included in \textit{MultiCaption} are derived from publicly available fact-checking articles. The released artifacts will include documentation describing dataset structure, labeling methodology, and intended use to facilitate responsible and reproducible research.

\paragraph{FAIR Principles.}
The release of \textit{MultiCaption} follows the FAIR data principles. The dataset will be \textit{Findable} through a persistent repository link provided upon acceptance, \textit{Accessible} under clearly defined research-only access conditions, \textit{Interoperable} via standard, machine-readable formats, and \textit{Reusable} through accompanying documentation describing dataset structure, labeling strategies, preprocessing steps, and intended use.

\end{document}